\documentclass[letterpaper, 10 pt, conference]{ieeeconf}

\IEEEoverridecommandlockouts
\usepackage{multicol,float}
\usepackage{flushend}
\usepackage{graphicx}
\usepackage{caption}
\usepackage[labelformat=simple]{subcaption}
\DeclareCaptionLabelSeparator{periodspace}{.\quad}
\makeatletter
\newcommand{\removelatexerror}{\let\@latex@error\@gobble}
\makeatother

\usepackage[lined,ruled,linesnumbered]{algorithm2e}
\usepackage[english]{babel}
\usepackage{blindtext}

\usepackage{amsmath} 
\usepackage{amssymb}  
\usepackage{todonotes}
\usepackage{pifont}
\usepackage{array}
\usepackage{centernot}

\usepackage[outdir=./figtmp/]{epstopdf}
\usepackage{xcolor}
\usepackage{pgf}
\pgfdeclarelayer{edgelayer}
\pgfdeclarelayer{nodelayer}
\pgfsetlayers{edgelayer,nodelayer,main}
\usepackage{tikz}
\tikzset{fontscale/.style = {font=\relsize{#1}}}
\usetikzlibrary{arrows, shapes, snakes, automata,
  backgrounds, positioning, calc, patterns, intersections}
\usepackage{url}
\usepackage{cite}
\usepackage{relsize}
\usepackage{color}
\usepackage{import}
\usepackage{comment}

\long\def\symbolfootnote[#1]#2{\begingroup%
  \def\thefootnote{\fnsymbol{footnote}}\footnote[#1]{#2}\endgroup}

\makeatletter
\newcommand\footnoteref[1]{\protected@xdef\@thefnmark{\ref{#1}}\@footnotemark}
\makeatother

\title{\LARGE \bf
Rope through Loop Insertion for Robotic Knotting:\\
A Virtual Magnetic Field Formulation
}

\author{Alejandro Marzinotto and Johannes A. Stork
\thanks{%
 The authors are with the 
Computer Vision and Active Perception Lab., 
Centre for Autonomous Systems, 
School of Computer Science and Communication, 
KTH Royal Institute of Technology, SE-100 44 Stockholm, Sweden.
e-mail: \tt{ $\{\,$almc$\,|\,$jastork$\,\}$@kth.se}}}

\begin{document}
\maketitle
\thispagestyle{empty}
\pagestyle{empty}

\begin{abstract}

Inserting an end of a rope through a loop is a common and important action that is required for creating most types of knots. To perform this action, we need to pass the end of the rope through an area that is enclosed by another segment of rope.
As for all knotting actions, the robot must for this exercise control over a semi-compliant and flexible body whose complex 3d shape is difficult to perceive and follow. Additionally, the target loop often deforms during the insertion.

We address this problem by defining a virtual magnetic field through the loop's interior and use the Biot Savart law to guide the robotic manipulator that holds the end of the rope.
This approach directly defines, for any manipulator position, a motion vector that results in a path that passes through the loop. The motion vector is directly derived from the position of the loop and changes as soon as it moves or deforms.

In simulation, we test the insertion action against dynamic loop deformation of different intensity. We also combine insertion with grasp and release actions, coordinated by a hybrid control system, to tie knots in simulation and with a NAO robot.



\end{abstract}

\section{Introduction}
\label{sec:introduction}

\begin{figure}[t]
\centering
\includegraphics[width=0.85 \linewidth]{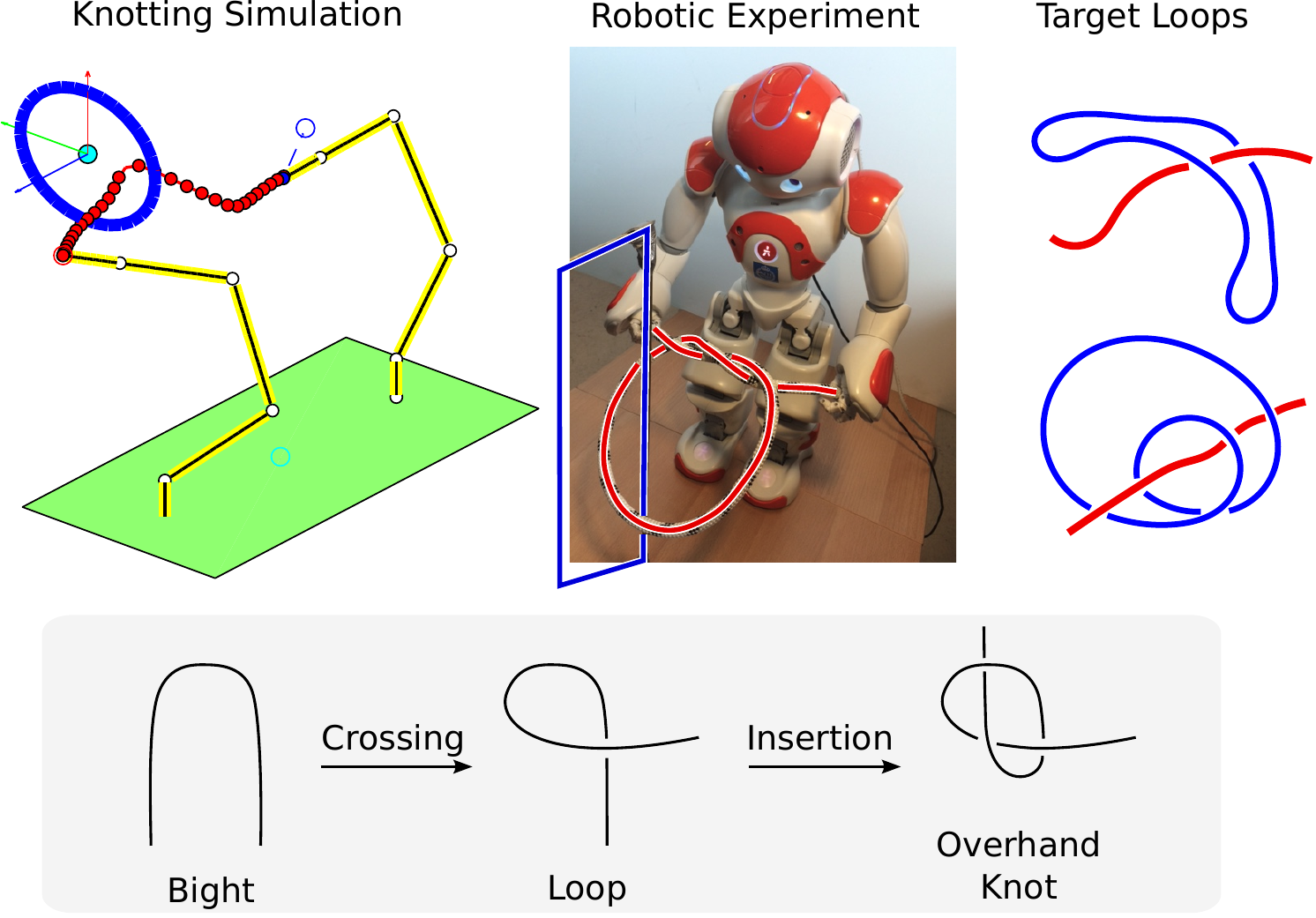}
\caption{%
\emph{Left:} Knotting simulation with robot's arms shown in yellow, robot's base in green, deformable rope in red, and target loop in blue.
\emph{Middle:} Experiment with NAO robot.
\emph{Right:} The target loops can have complex three-dimensional shapes.
\emph{Bottom:} Knot as a sequence of basic actions.
}
  \label{fig:overview}
  \vspace{-0.5cm}
\end{figure}

Making knots with a length of rope is a skill that benefits humans in many everyday activities and there exists a large variety of knots for different purposes~\cite{knotsguide, ashley1993}.
Every knot is created by a sequence of knotting actions: For example, imagine a straight length of rope. Folding it to a bight so that both ends lie alongside each other. Crossing one end over the other results in a loop. Pushing the other end through the loop and pulling on both ends creates the \emph{overhand} type knot shown in the bottom of Fig.~\ref{fig:overview}.

What makes knotting difficult for robots is that it requires dexterous manipulation of a semi-compliant, flexible body in form of bending, twisting, holding, and inserting, as in the example above. Therefore, knotting often requires coordinating two or more manipulators and feedback control.

%

In this work, we focus on \emph{insertion through a target loop}. This is the most complex of the basic knotting actions as it requires moving a part of the rope, e.g. an end of the rope, on a trajectory relative to another segment of rope that forms the loop. 
For this purpose we abstract the situation of the target loop with its crossing and two adjacent rope parts, shown at the bottom middle of Fig.~\ref{fig:overview}, and instead consider the closed target loop and the length of rope to be inserted separately, as seen in blue and red in the knotting simulation of Fig.~\ref{fig:overview}.

In three-dimensional space, the target loop can assume complex shapes, such as the ones displayed on the right in Fig.~\ref{fig:overview}. This makes \emph{ad hoc} definitions of a center point or an inner surface for the loop---as they would be possible for planar loops---futile. This variety in loop shapes consequently prohibits simple geometric 
parameterization of the required manipulator motion for insertion---such as for instance just approaching the loop's center or crossing the loop's inner surface from one side to the other.
%

On the contrary, to guide the manipulator holding the rope's end through the loop---and thus making sure that the rope gets tucked through the loop---we need to consider the spatial relationship of the trajectory and the loop. For any starting position of the manipulator, the trajectory must approach the loop, pass through, and leave on the other side. These conditions are fulfilled by all field lines of a magnetic field that is induced by an electric current flowing through the loop. Therefore, we model the insertion motion of the manipulator as that of a particle following the direction of the loop's \emph{virtual} magnetic force field.

The contributions are summarized as follows:
\begin{itemize}

\item Formulating insertion as a position-based controller with termination conditions---both based on the virtual magnetic field and applying to all shapes of 3d loops.

\item Coordinating basic knotting actions with a hybrid control system for knot-tying.

\item Evaluating insertion reliability with respect to loop deformation from simulated position noise.

\item Demonstrating the knotting system for a family of mathematical knots in simulation and with a NAO humanoid robot.

\end{itemize}



\section{Related Work}
\label{sec:related}

Inserting an end of a rope through a loop of rope or an anchoring entity, as in our insertion action, requires moving a manipulator into a hollow. In robotics, such tasks have been studied before: 
Most similar to our approach is the high precision threading framework in \cite{wangonline}. Similar to us, they exploit the insight that magnetic field lines pass through the interior of a current-carrying coil. However, instead of formulating a control task and computing derivatives as in \cite{wangonline}, we directly refer to the magnetic field's direction to define reference positions. While they consider a surgical context with tight openings modelled by small, planar loops that are circular and static, we consider rope-based knot-making with dynamically deforming loops of arbitrary three-dimensional shape. For the planar case we additionally introduce a parametrization that allows adapting the manipulator's trajectory. 

Further similar works based on topological coordinates are found in motion planning \cite{ivan2013topology, Ho:ICRA2013, Ho:TopologyIK:PG09, Ho_Komura:CAVW2011} and caging \cite{pokorny2013holes, stork2013integrated}. Instead of workspace coordinates, as in our system, these works refer to coordinates in terms of writhe \cite{ivan2013topology, Ho:ICRA2013, Ho:TopologyIK:PG09, Ho_Komura:CAVW2011}, winding \cite{pokorny2013holes}, or linking numbers \cite{stork2013integrated} and define success as reaching goal coordinates. In contrast, we define switching conditions for the insertion action as passing the magnetic field's maximum intensity. These works employ approximate inference \cite{ivan2013topology}, optimization \cite{Ho:ICRA2013, Ho:TopologyIK:PG09, Ho_Komura:CAVW2011, pokorny2013holes}, or random-based planning \cite{stork2013integrated} to generate manipulator trajectories. However, we directly refer to a virtual force field to derive the direction of motion. While small changes in workspace position can lead to fundamental changes in topological coordinates, we show that our control based on magnetic fields is robust to position noise.

In this work, we use insertion of an end of a rope as a central basic action in conjunction with other basic actions to create knots. Several other works also abstract the knotting problem by subdividing it into a set of basic steps\cite{matsuno2006manipulation, yamakawa2008knotting, van2012new, kudoh2015inair, wakamatsu2006knotting}. Often these works consider mathematical knots and implement Reidemeister moves that locally change the knot's link diagram\cite{matsuno2006manipulation, yamakawa2008knotting, van2012new, kudoh2015inair, wakamatsu2006knotting}. In contrast, we define a set of five basic actions---including grasping and releasing of the rope, twisting the rope, and inserting the rope through a loop---to create knots. Instead of representing the knot by its linking diagram, we note the sequence of actions that are required for its creation, for instance in form for a Behavior Tree~\cite{ale_icra2013}.

Speed and repeatability for industrial robotic knotting have been studied \cite{kenji1975automatic, bell2014knot}, often referring to feed forward control for simple knots~\cite{yamakawa2010motion, kudoh2015inair}. 
Most of this research, formulates knotting as a motion planning problem in the context of deformable object manipulation~\cite{saha2006motion, saha2007manipulation, hashimoto2002dynamic, yamakawa2010motion, matsuno2001flexible, lynch1999dynamic}. In contrast, we do not model the rope or its physical properties. Instead, we assume that the rope is held by a manipulator and will comply to its motion when twisted and tucked through a loop. 

All above motion planning approaches \cite{saha2006motion, saha2007manipulation, hashimoto2002dynamic, yamakawa2010motion, matsuno2001flexible, lynch1999dynamic, Ho:ICRA2013, Ho:TopologyIK:PG09, Ho_Komura:CAVW2011, stork2013integrated, ivan2013topology} have in common that they require time intensive re-planning if the situation changes. However, in knotting we manipulate a flexible body and move on trajectories defined with respect to other parts of the rope which constantly causes dynamic changes. We address this problem by employing feedback control that is directly based on the loop's shape and position for the insertion action.

\section{Virtual Magnetic Field-based Rope Insertion}
\label{sec:insertion-control}

To insert an end of a rope that is held by a manipulator though a loop, $\mathcal{L}$, we need to move the manipulator on a trajectory that passes through the loop's hollow. In our approach, we generate such a trajectory by continuously feeding reference positions for the manipulator, $\mathbf{x}^{r} \in \mathbb{R}^3$, to the manipulator's controller. In each time step, $t$, we compute the reference position from the manipulator's current position, $\mathbf{x}^{c}(t)$, and a directional offset, $\boldsymbol \delta(t)$, 
such that 
$
\mathbf{x}^{r} (t) 
=
\mathbf{x}^{c}(t) + \boldsymbol \delta(t)
$.
%

Below, we define the directional offset, $\boldsymbol \delta(t)$, based on the magnetic field generated by a steady electric current, $I$, through the loop, $\mathcal{L}$. The definition relates direction, length, and proximity of current in $\mathcal{L}$ to $\boldsymbol \delta(t)$ and gives the direction in which the manipulator has to move. For this, we adopt the Biot Savart law and model $\mathcal{L}$ as an infinitely-narrow, closed wire.

\subsection{Adapting the Biot Savart Law for Control}
\label{sec:biotsavart}

The Biot Savart law relates the magnetic field, $\mathbf{B}$, at any coordinate, $\mathbf{x} \in \mathbb{R}^3$, to the electric current in the conductor $\mathcal{L}$. In Eq.~\eqref{eq:biot_savart}, it is expressed as a line integral, exploiting the superposition principle for magnetic fields.
\begin{equation}
\mathbf{B}(\mathbf{x}) 
= 
\frac{\mu_0}{4\pi} 
I\,
\int_{\mathcal{L}} 
\frac{%
\mathrm{d}\mathbf{l} \times (\mathbf{x} - \mathbf{x}')
}{%
\|\mathbf{x} - \mathbf{x}'\|^3
}
.
\label{eq:biot_savart}
\end{equation}
Above, $\mu_0$ is the magnetic constant; $I$ is current intensity; $\mathrm{d}\mathbf{l}$ is the length differential of the conductor $\mathcal{L}$; $\mathbf{x}'$ is the position of a given segment $\mathrm{d}\mathbf{l}$; and $\mathbf{x} - \mathbf{x}'$ is a vector from $\mathrm{d}\mathbf{l}$ to $\mathbf{x}$. In case $\mathcal{L}$ is approximated by $n$ line segments, Eq.~\eqref{eq:biot_savart} reduces to a sum, allowing computation of $\mathbf{B}(\mathbf{x})$ with time complexity $\mathcal{O}(n)$.

The magnetic flux density, as given by Eq.~\eqref{eq:biot_savart}, drops with increasing distance from the conductor,  $\mathbf{x} - \mathbf{x}'$. To guarantee that our insertion action has constant and adjustable approach speed, we normalize the magnetic field when defining the directional offset,
\begin{equation}
\boldsymbol \delta(t)
=
\gamma \,
\mathbf{B}(\mathbf{x}) 
/ 
\| \mathbf{B}(\mathbf{x}) \|
\label{eq:biot_savart_normal}
.
\end{equation}
In this, we set the coordinate to the manipulator's current position, $\mathbf{x} = \mathbf{x}^{c}(t)$, and use $\gamma$ as a scaling factor. 

\begin{figure}[h]
\centering
\includegraphics[width=0.6\linewidth, trim={0cm 0cm 0cm 1cm}, clip]{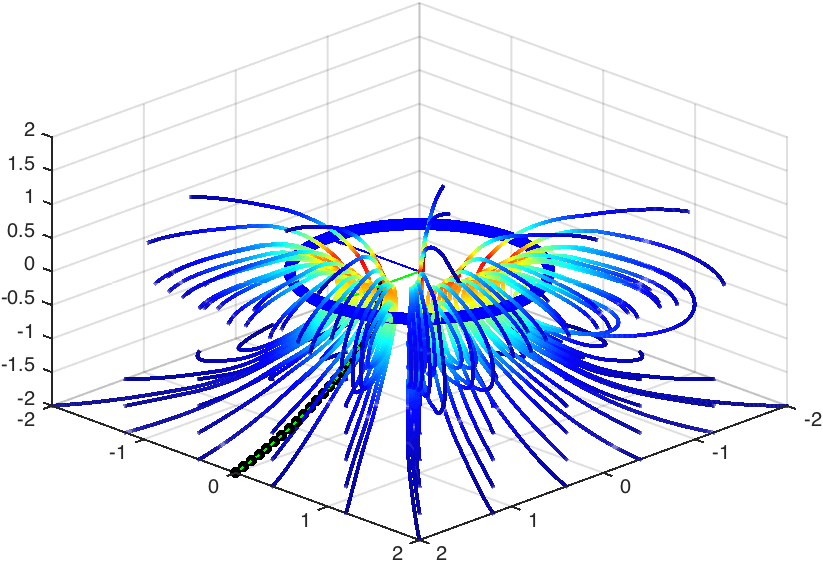}
\vspace*{0.1cm}
\includegraphics[width=0.6\linewidth, trim={0cm 0cm 0cm 1cm}, clip]{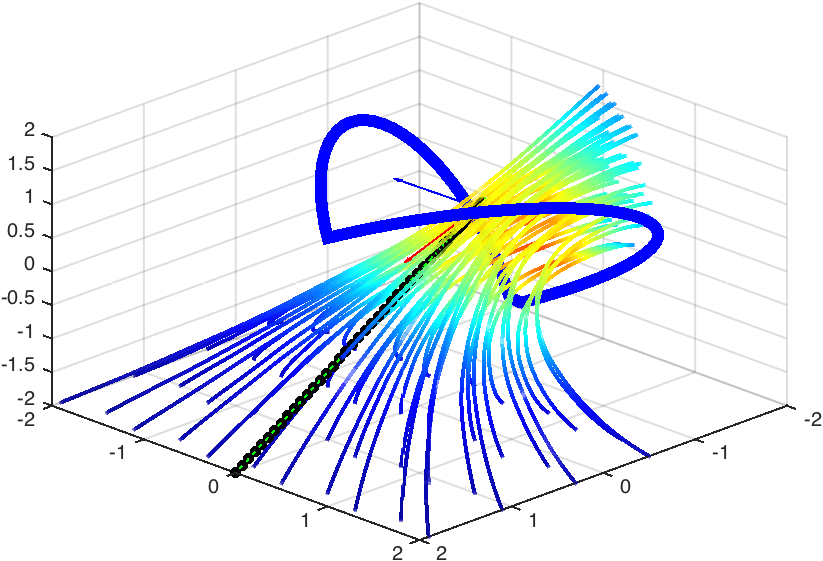}
\vspace*{0.1cm}
\includegraphics[width=0.6\linewidth, trim={0cm 0cm 0cm 1cm}, clip]{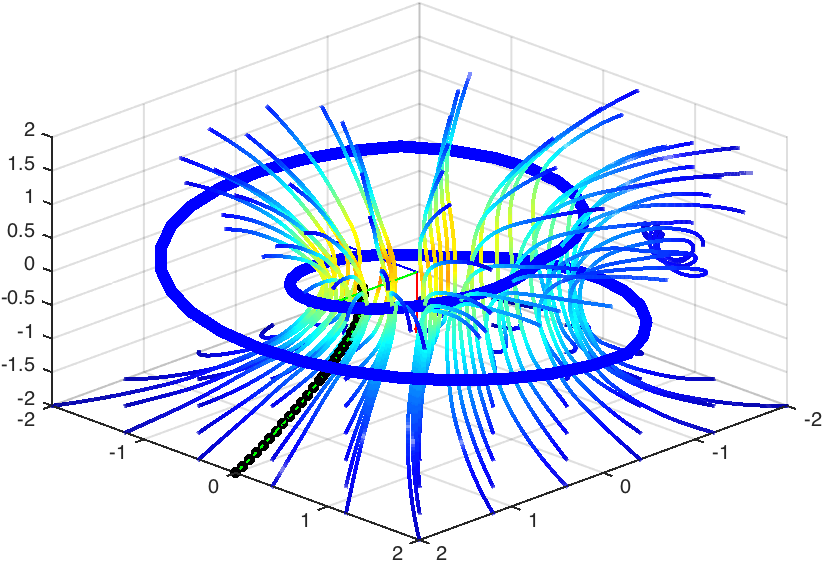}

\caption{%
Magnetic fields for loops of several shapes illustrated by the field lines. 
Color indicates magnetic flux density, i.e. the intensity of the magnetic field.
Tracking a field line until the field drops off (black line) stops the manipulator in the ``middle" of the loop.}
\label{fig:field-lines}
\label{fig:stopping}
\end{figure}

As seen in Fig.~\ref{fig:field-lines}, every field line passes through the loop by entering the electromagnet near its south pole and exiting near its north pole. This allows us to decide from which side the insertion is performed by altering the direction of electric current in $\mathcal{L}$. Since any given coordinate, $\mathbf{x}$, lies on some field line, we can start with the manipulator at any position and pass through the loop by following $\mathbf{B}$. Importantly, we can identify that the manipulator has passed through the loop when the flux density, visualized in Fig.~\ref{fig:field-lines} by color, ceases to increase and starts to decrease.

\subsection{Parametrization for Planar Loops}

\begin{figure}[]
  \centering
  \begin{subfigure}[b]{0.45\linewidth}
	\includegraphics[width=\textwidth,trim={5cm 5cm 5cm 3.3cm},clip]{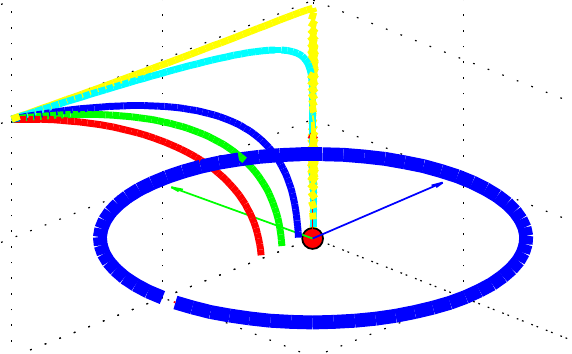}
	
    \caption{\footnotesize red: $\alpha = \beta$, green: $\alpha = \sqrt{2}\beta$, blue: $\alpha = 2\beta$, cyan: $\alpha = 10\beta$, yellow: $\alpha = 100\beta$.}
    
    \label{fig:alpha_g_beta}
  \end{subfigure}%
  \hspace{10pt}
  \begin{subfigure}[b]{0.45\linewidth}
    \includegraphics[width=\textwidth,trim={5cm 5cm 5cm 3.3cm},clip]{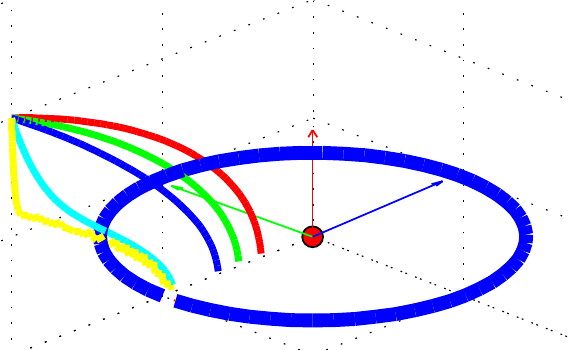}
    
    \caption{\footnotesize red: $\alpha = \beta$, green: $\alpha = \frac{1}{\sqrt{2}}\beta$, blue: $\alpha = \frac{1}{2}\beta$, cyan: $\alpha = \frac{1}{10}\beta$, yellow: $\alpha = \frac{1}{100}\beta$.}
    
    \label{fig:alpha_l_beta}
    \end{subfigure}
  \caption{Insertion trajectory for planar loops parametrized by $\alpha$ and $\beta$. 
  }
  \label{fig:alpha_beta}
\end{figure}

Many knots are described by a two-dimensional layout, similar to the diagrams in Fig.~\ref{fig:overview}, using the 3rd dimension merely for crossings of rope segments~\cite{knotsguide, ashley1993}. Therefore, the insertion action is often performed on mostly planar loops. Below, we give a parameterization of the magnetic field for planar loops that allows changing the shape of the insertion trajectory, as exemplified in Fig.~\ref{fig:alpha_beta}, by adjusting two scalar parameters, $\alpha$ and $\beta$. We can, for example, continuously deform a trajectory if the computed insertion trajectory  is obstructed.

\begin{figure}
\centering 
\hspace*{1cm}\includegraphics[width=0.7\textwidth, trim={2.5cm 13.7cm 5cm 9.2cm},clip]{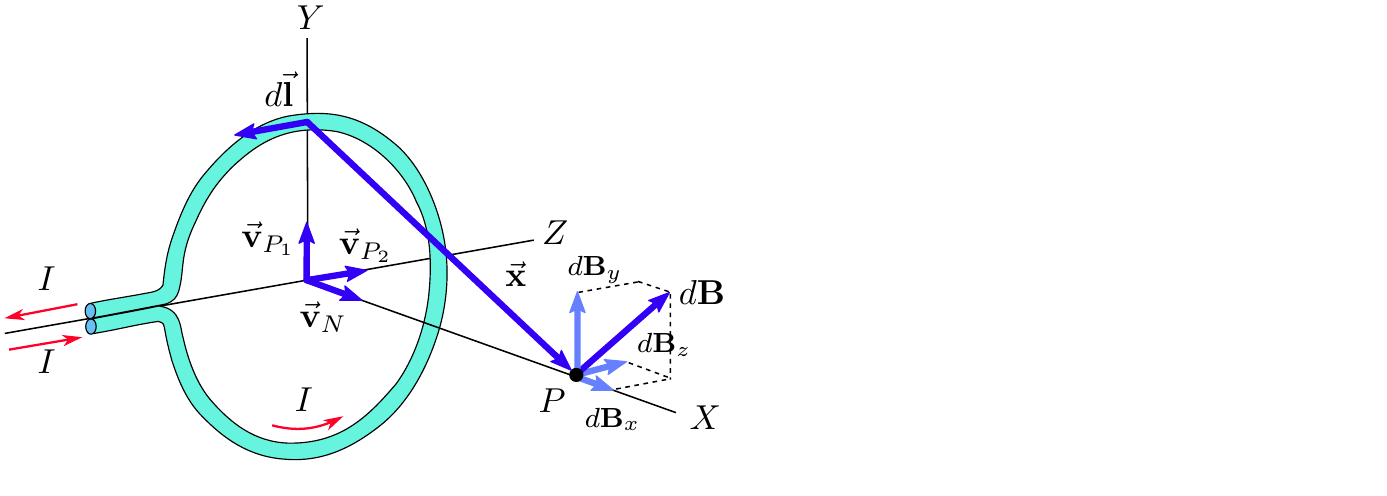}
\caption{%
Parameterization for a planar conductor (cyan). Image adapted from, Pearson Education, Inc. publishing as Addison Wesley. 
\protect\url{http://www.pearsoned.co.uk/imprints/addison-wesley/}
}
\label{fig:biot_savart2}
\end{figure}

A planar loop lies on a plane that defines one normal and two parallel vectors to the loop: $\mathbf{v}_N$, $\mathbf{v}_{P_1}$ and $\mathbf{v}_{P_2}$. In Fig.~\ref{fig:biot_savart2}, these three vectors correspond to the axes $X$, $Y$ and $Z$, respectively. 
We project $\mathbf{B}(\mathbf{x})$ into this coordinate system to represent the magnetic field in terms of two scalar components, $B_{P_1}$ and $B_{P_2}$, which are parallel to the loop, and one scalar component, $B_{N}$, which is perpendicular to it. This yields the following expression for the magnetic field:
\begin{equation}
\mathbf{B}(\mathbf{x}) 
=  
B_{P_1}(\mathbf{x})\mathbf{v}_{P_1} + B_{P_2}(\mathbf{x})\mathbf{v}_{P_2} + B_N(\mathbf{x})\mathbf{v}_N
\label{eq:biot_savart_separated}
.
\end{equation}
We introduce the parameters  $\alpha$ and $\beta$  into Eq.~\eqref{eq:biot_savart_separated} to control the weight of the parallel and perpendicular components of the magnetic field,
\begin{equation}
\mathbf{B}(\mathbf{x}, \alpha, \beta) 
= 
\alpha B_{P_1}(\mathbf{x})\mathbf{v}_{P_1} 
+
\alpha B_{P_2}(\mathbf{x})\mathbf{v}_{P_2}
+ 
\beta B_N(\mathbf{x})\mathbf{v}_N
\label{eq:biot_savart_param}
\end{equation}
and define the directional offset analogous to Eq.~\eqref{eq:biot_savart_normal},
\begin{equation}
\boldsymbol \delta(t, \alpha, \beta)
=
\gamma \,
\mathbf{B}(\mathbf{x}, \alpha, \beta) 
/ 
\| \mathbf{B}(\mathbf{x}, \alpha, \beta) \|
\label{eq:biot_savart_normal_parameters}
.
\end{equation}
The influence of $\alpha$ and $\beta$ on the manipulator's trajectory is exemplified in Fig.~\ref{fig:alpha_beta}.

\section{Knotting System Overview}
\label{sec:framework}

As illustrated in the example of Sec.~\ref{sec:introduction}, making a knot consists in performing several basic knotting actions of which the insertion action is central. Since the rope needs to be handed over after insertion, we consider a two-armed robot with a movable base (as in Fig.~\ref{fig:overview}). We formulate the knotting system as two feedback controllers---one for the base and one for the active manipulator---that are feed with reference points for each time step and action. The knotting process is governed by a hybrid system that decides on reference points and switching conditions 
which determine when the next action needs to be triggered.
We present examples of hybrid systems for different knots in Sec.~\ref{sec:technique}. In this section, we detail how we realize individual basic knotting actions and define their switching conditions. 

\subsection{Basic Knotting Actions}
\label{ssec:action_primitives}

\paragraph{Grasp Rope}
\label{sssec:grasp_rope}
Set the manipulator's reference to point on the rope; close gripper when reached. 
\emph{Switching:} target point reached and gripper closed.

\paragraph{Release Rope}
\label{sssec:release_rope}
Set the manipulator's reference to same point from previous time step; open gripper. 
\emph{Switching:} gripper opened.

\paragraph{Twist Rope}
\label{sssec:twist_rope}
Set the references of both manipulators to a predefined circular trajectory such that a loop is formed. 
\emph{Switching:} trajectory completed.

\paragraph{Turn Base}
\label{sssec:turn_base}
Set the base's reference to rotate the robot about itself such that the manipulators twist the rope. 
\emph{Switching:} trajectory completed.

The last two actions are redundant. In Sec.~\ref{sec:technique} we exploit this to show that knotting action sequences are compliant with the type of plans Behavior Trees (BTs)~\cite{ale_icra2013} represent.

\paragraph{Insertion}
\label{sssec:insertion}

Set the references of both manipulators using the target loop and the (parametrized) magnetic field in Sec.~\ref{sec:insertion-control} such that they approach the loop from opposite sides. 
%
%
\emph{Switching}: the magnetic field drops when following the trajectory further. 

The insertion action requires combination with a grasp and a release action to thread the rope through the loop.



\section{Tying Knots}
\label{sec:technique}

\begin{figure}[h]
  \includegraphics[width=\linewidth, trim={0 6.7cm 0 0}, clip]{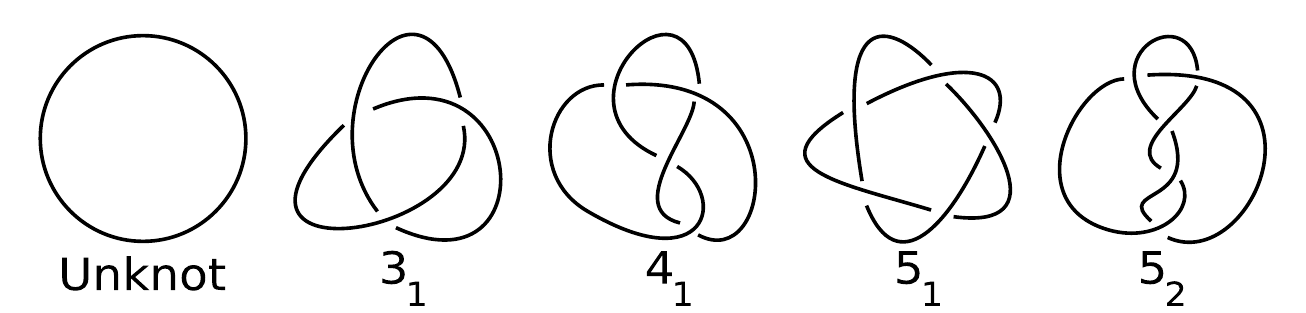}
  \caption{A family of mathematical knots that can be tied using only two manipulators and their designation in A-B notation. Knot $3_1$ is called \emph{trefoil}.}
  \label{fig:overview-table}
\end{figure}

In this section, we describe hybrid control systems, that govern the knotting process and schedule basic knotting actions from Sec.~\ref{sec:framework}. For this, we first refer to the family of mathematical knots shown in Fig.~\ref{fig:overview-table}, since they can be tied with only two manipulators. In Sec.~\ref{ssec:beyond} we consider other types of knots based on the same knotting actions.\footnote{Knotting system and simulation: \url{www.github.com/~almc}.}

\subsection{The Trefoil Knot}
\label{ssec:trefoil_knot}

To tie the right-handed trefoil knot---referred to as $3_1$ in Fig.~\ref{fig:overview-table}---we require 9 steps (step 2 to 10) of which two use insertion actions. A step consists of one basic action instantiated within a particular context. When formulating the action sequence, we refer to the two ends of the rope as $R_0$ and $R_f$. Further, we assume that the rope has reasonable internal stiffness, the target loops are large enough for insertion with the manipulators, and that the manipulators can grasp and hold the rope. To support the rope, we employ to an \emph{anchoring entity} in front of the robot. An illustration of all steps listed below is given in Fig.~\ref{fig:knotting_all_steps}.



\subsubsection{Approach Loop}
Use base controller with reference to loop position and orientation. (Not a basic action.)



\subsubsection{Arm~2 Grasps $R_f$}
Using \emph{Grasp Rope}.



\subsubsection{Arm~1,2 Insert $R_f$ Through Anchoring Entity}
\label{sssec:p2}
Use \emph{Insertion}.



\subsubsection{Arm~1 Grasps $R_f$, Arm~2 Releases $R_f$}
Use \emph{Grasp} and \emph{Release Rope}.



\subsubsection{Arm~2 Grasps $R_0$}
Use \emph{Grasp Rope}.



\subsubsection{Constructing Rope Loop}
\label{sssec:construct_rope_loop}
Use \emph{Turn Base} (6.1) or \emph{Twist Rope} (6.2).

\subsubsection{Arm~1 Passes $R_0$ Through Rope Loop}
Use \emph{Insertion}.



\subsubsection{Arm~2 Releases $R_f$}
Use \emph{Release Rope}.



\subsubsection{Arm~2 Grasps $R_0$, Arm~1 Releases $R_0$}
Use \emph{Grasp} and \emph{Release Rope}.



\subsubsection{Arm~1 Grasps $R_f$}
Use \emph{Grasp Rope}.

\begin{figure*}
        \centering
        \begin{subfigure}[b]{0.33\textwidth}
                \includegraphics[width=\textwidth, trim={0cm 0cm 0cm 2cm}, clip]{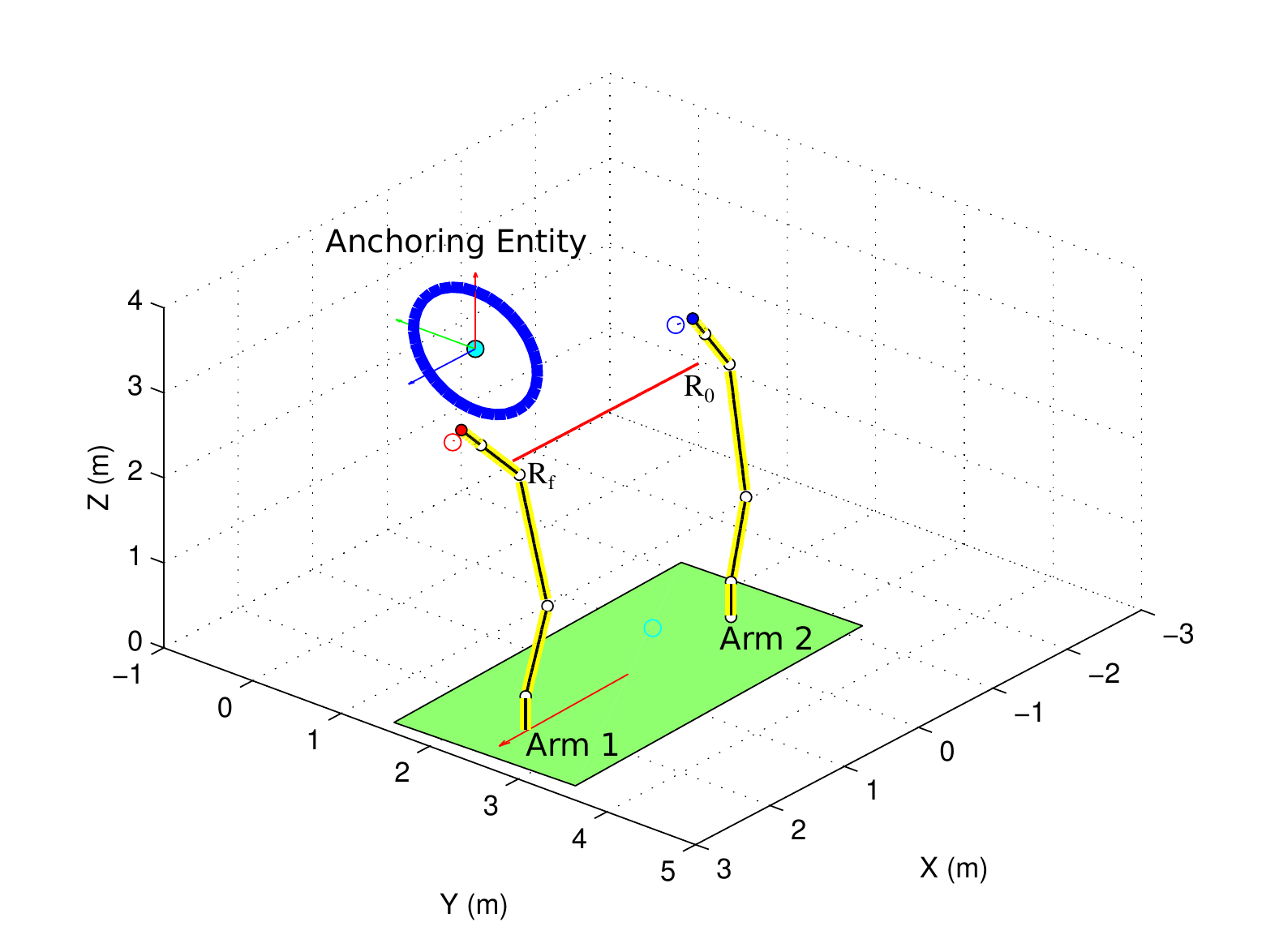}
                \vspace*{-0.7cm}
                \caption{Base Approaches Loop}
               \vspace*{0.1cm}
                \label{fig:p0}
        \end{subfigure}%
        \begin{subfigure}[b]{0.33\textwidth}
                \includegraphics[width=\textwidth, trim={0cm 0cm 0cm 2cm}, clip]{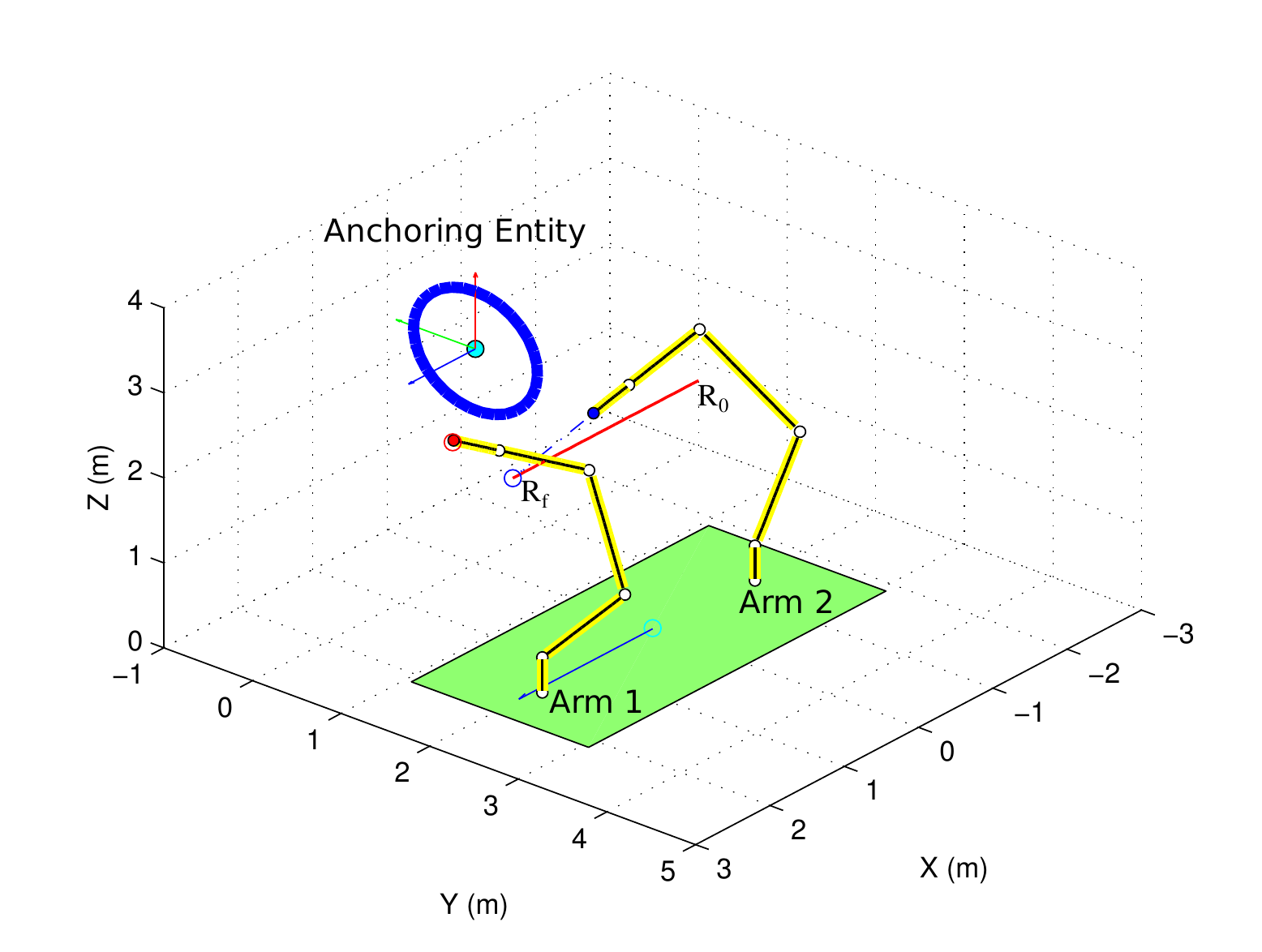}
                \vspace*{-0.7cm}
                \caption{Arm~2 Grasps $R_f$}
                \vspace*{0.1cm}
                \label{fig:p1}
        \end{subfigure}%
        \begin{subfigure}[b]{0.33\textwidth}
                \includegraphics[width=\textwidth, trim={0cm 0cm 0cm 2cm}, clip]{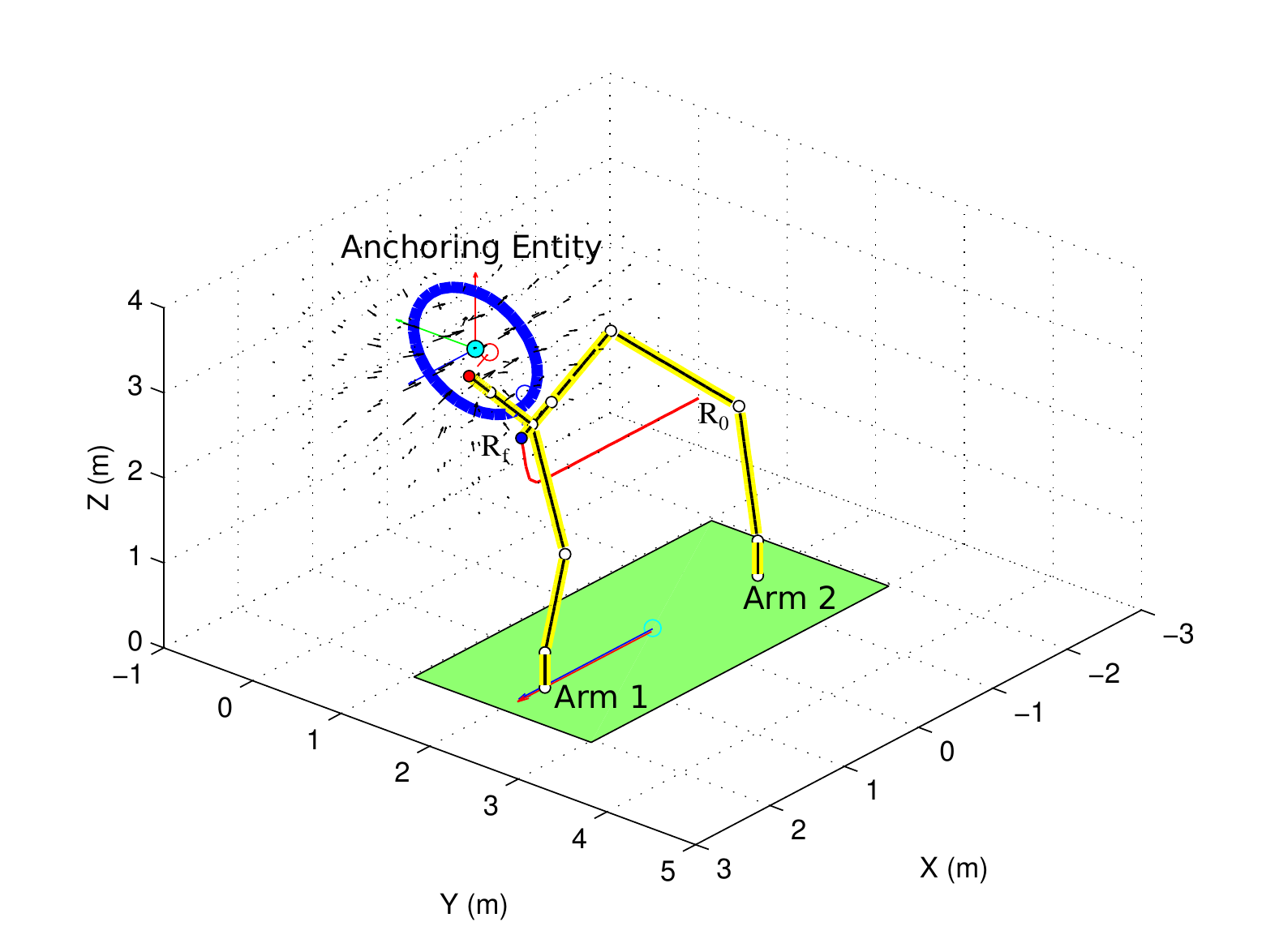}
                \vspace*{-0.7cm}
                \caption{Arm~1,2 Insert $R_f$ Through Anchoring Entity}
                \vspace*{0.1cm}
                \label{fig:p2}
        \end{subfigure}\\

        \begin{subfigure}[b]{0.33\textwidth}
                \includegraphics[width=\textwidth, trim={0cm 0cm 0cm 2cm}, clip]{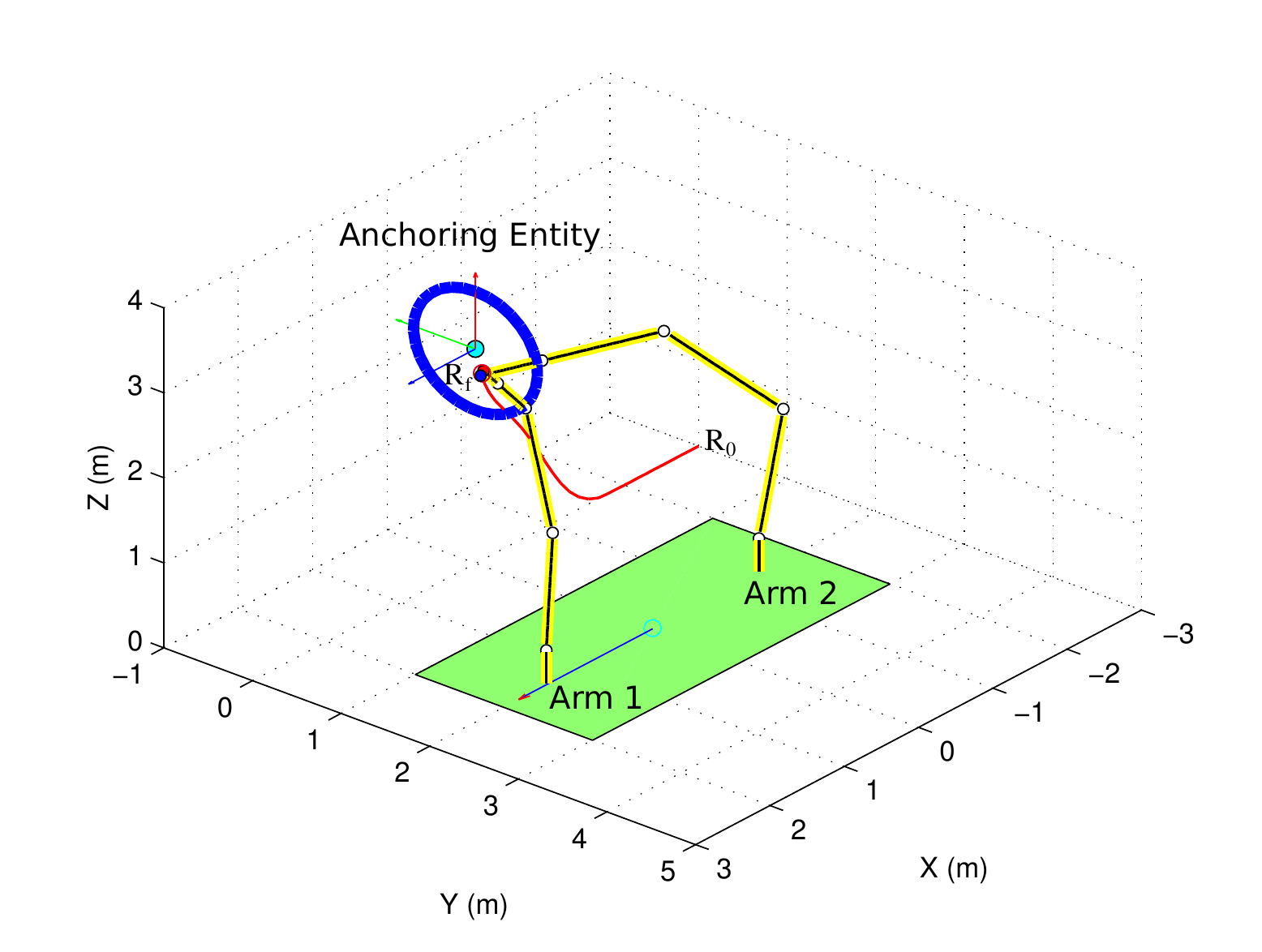}
                \vspace*{-0.7cm}
                \caption{Arm~1 Grasps $R_f$, Arm~2 Releases $R_f$}
                \vspace*{0.1cm}
                \label{fig:p3}
        \end{subfigure}%
        \begin{subfigure}[b]{0.33\textwidth}
                \includegraphics[width=\textwidth, trim={0cm 0cm 0cm 2cm}, clip]{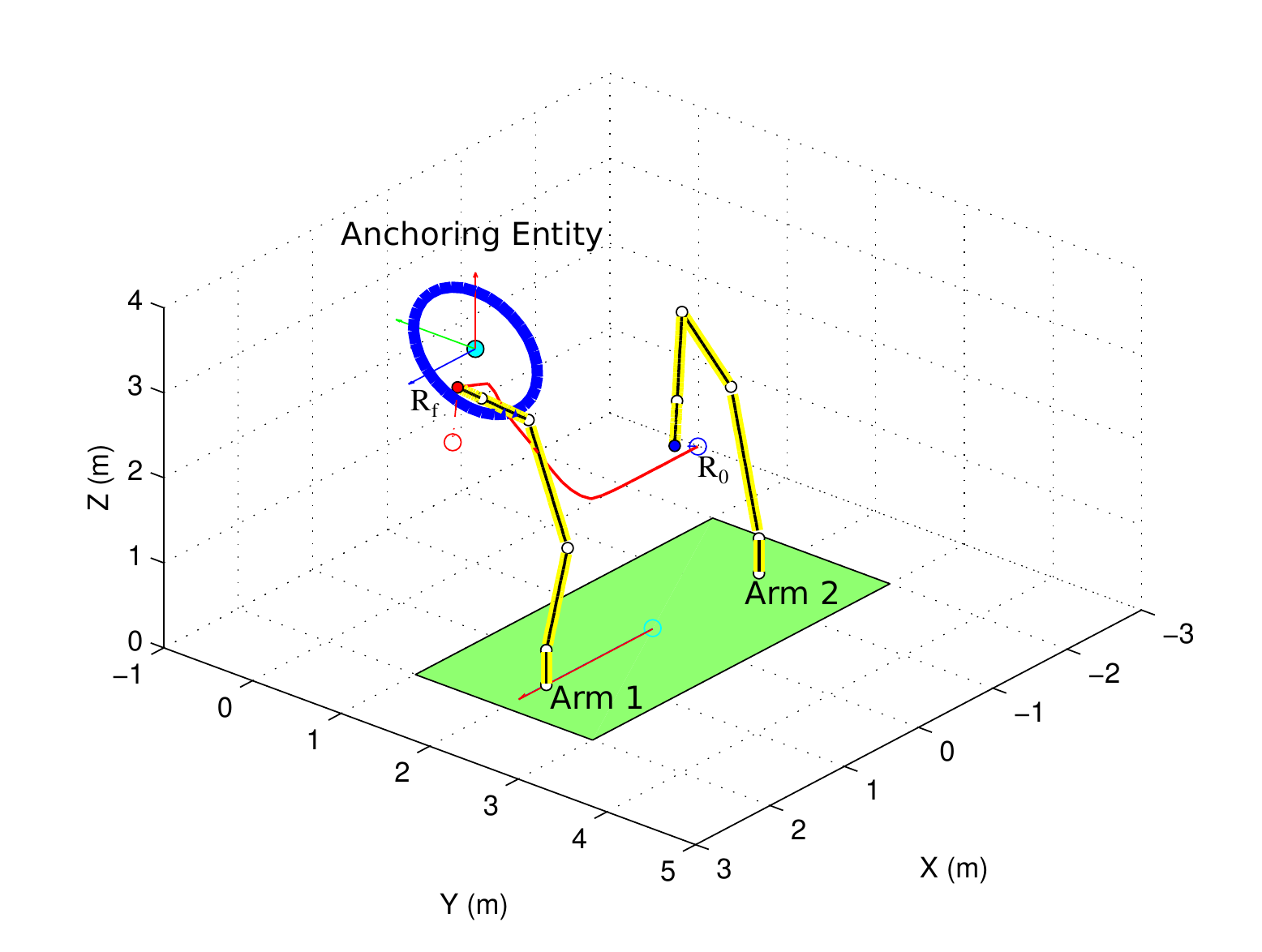}
                \vspace*{-0.7cm}
                \caption{Arm~2 Grasps $R_0$}
                \vspace*{0.1cm}
                \label{fig:p4}
        \end{subfigure}%
        \begin{subfigure}[b]{0.33\textwidth}
                \includegraphics[width=\textwidth, trim={0cm 0cm 0cm 2cm}, clip]{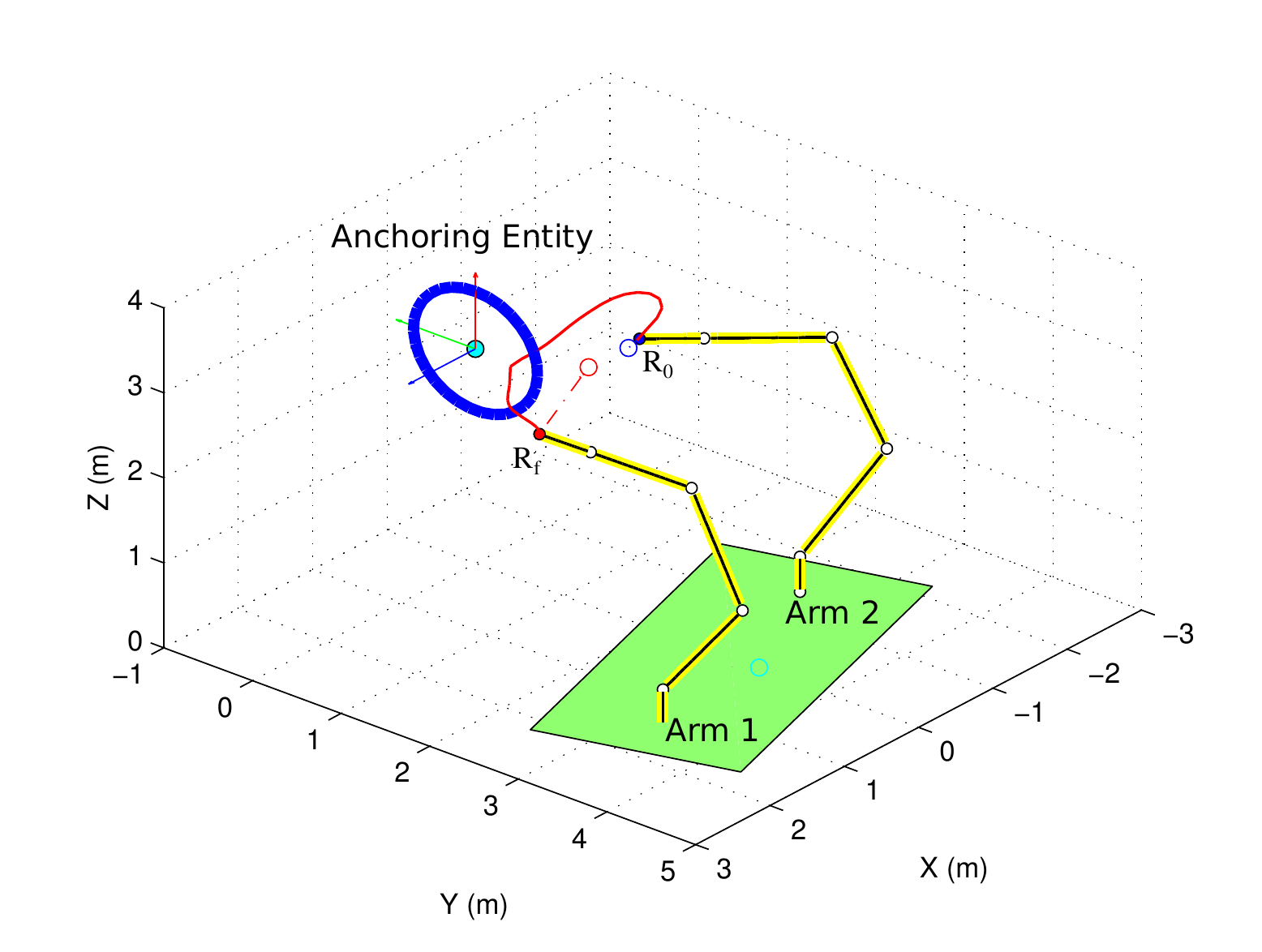}
                \vspace*{-0.7cm}
                \caption{Base Turns $\pi$ Around Itself (part 1)}
                \vspace*{0.1cm}
                \label{fig:p5a}
        \end{subfigure}\\

        \begin{subfigure}[b]{0.33\textwidth}
                \includegraphics[width=\textwidth, trim={0cm 0cm 0cm 2cm}, clip]{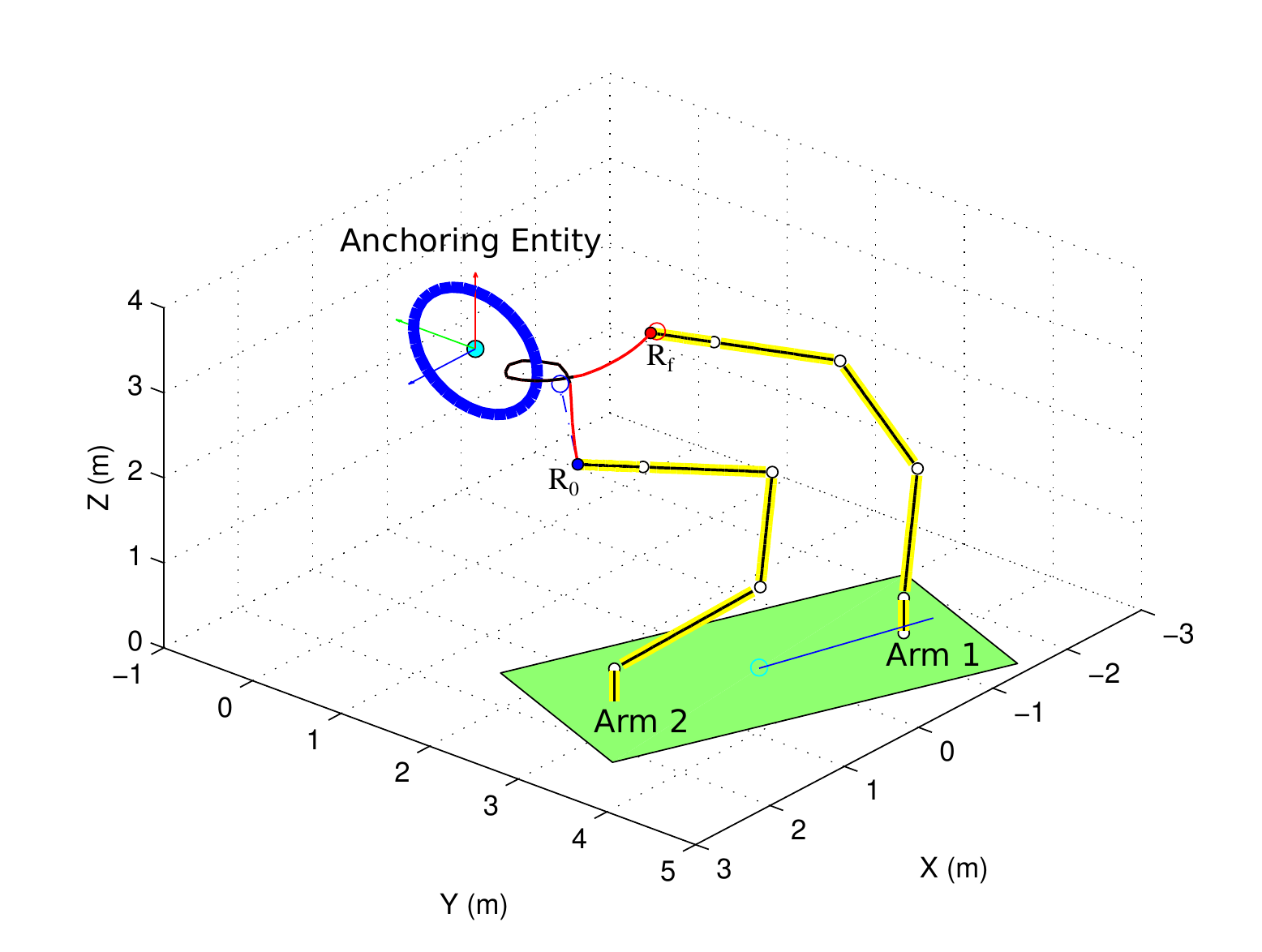}
                \vspace*{-0.7cm}
                \caption{Base Turns $\pi$ Around Itself (part 2)}
                \vspace*{0.1cm}
                \label{fig:p5b}
        \end{subfigure}%
        \begin{subfigure}[b]{0.33\textwidth}
                \includegraphics[width=\textwidth, trim={0cm 0cm 0cm 2cm}, clip]{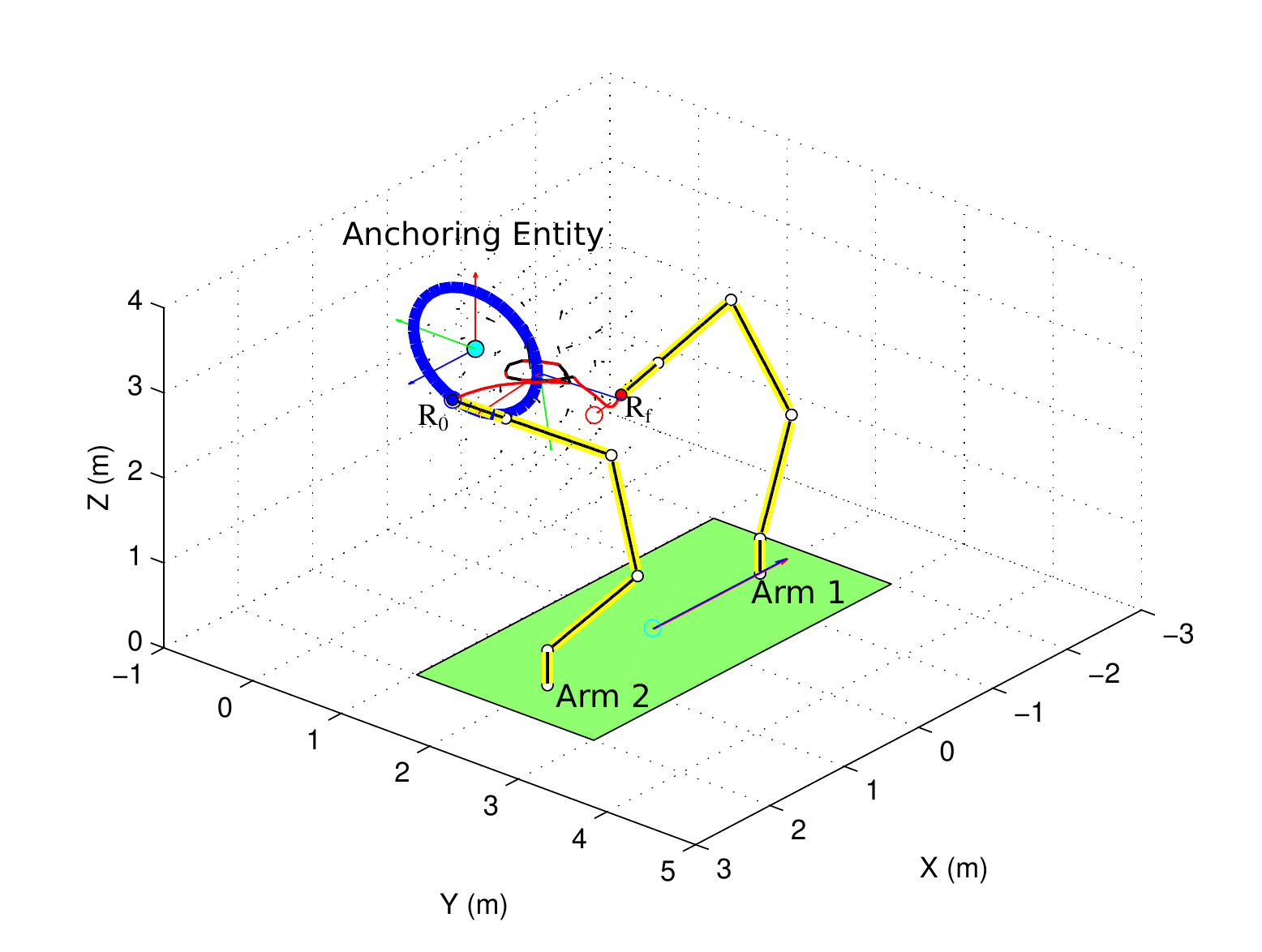}
                \vspace*{-0.7cm}
                \caption{Arm~1 Passes $R_0$ Through Rope Loop}
                \vspace*{0.1cm}
                \label{fig:p6}
        \end{subfigure}%
        \begin{subfigure}[b]{0.33\textwidth}
                \includegraphics[width=\textwidth, trim={0cm 0cm 0cm 2cm}, clip]{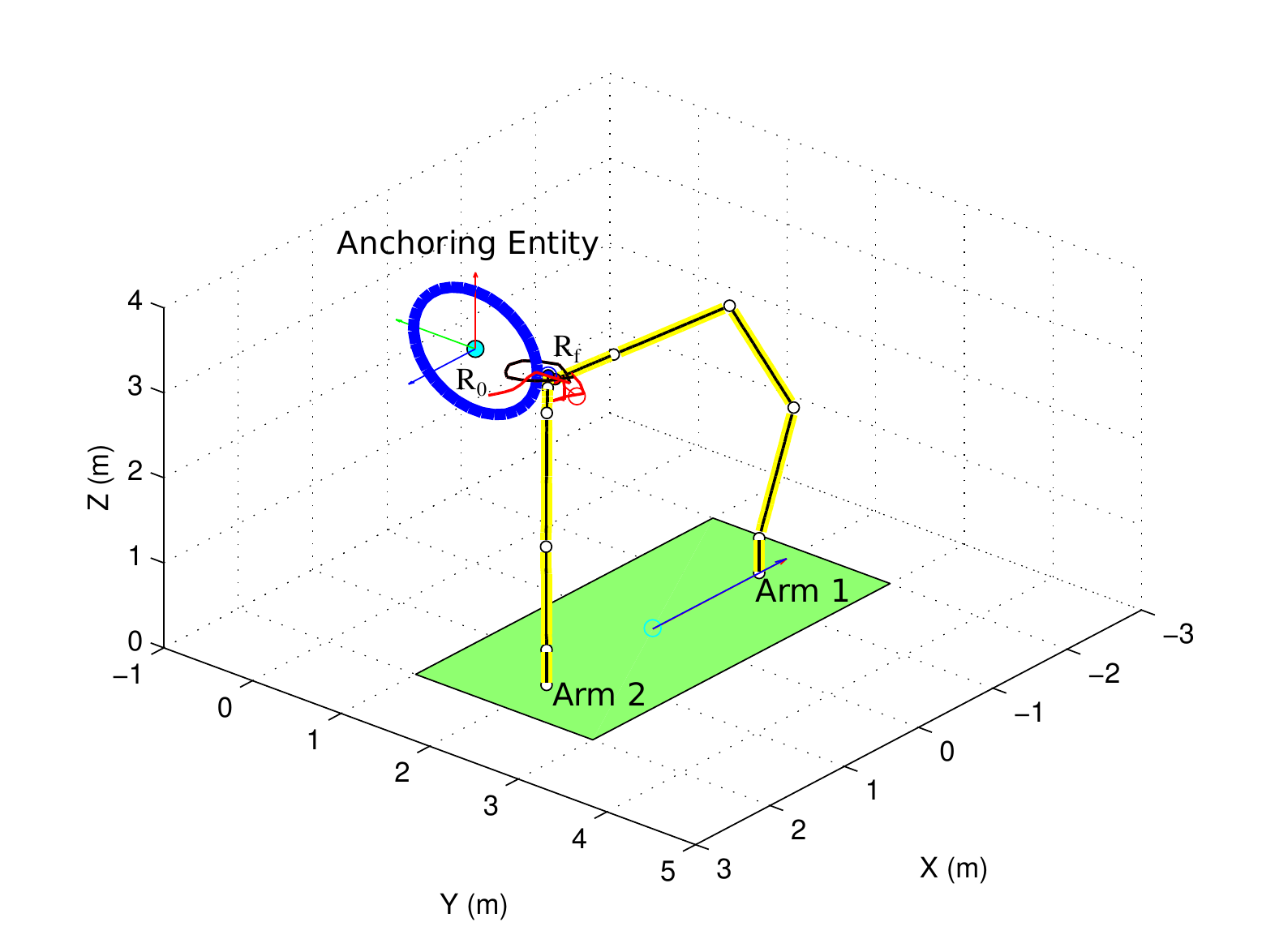}
                \vspace*{-0.7cm}
                \caption{Arm~2 Releases $R_f$}
                \vspace*{0.1cm}
                \label{fig:p7}
        \end{subfigure}\\

        \begin{subfigure}[b]{0.33\textwidth}
                \includegraphics[width=\textwidth, trim={0cm 0cm 0cm 2cm}, clip]{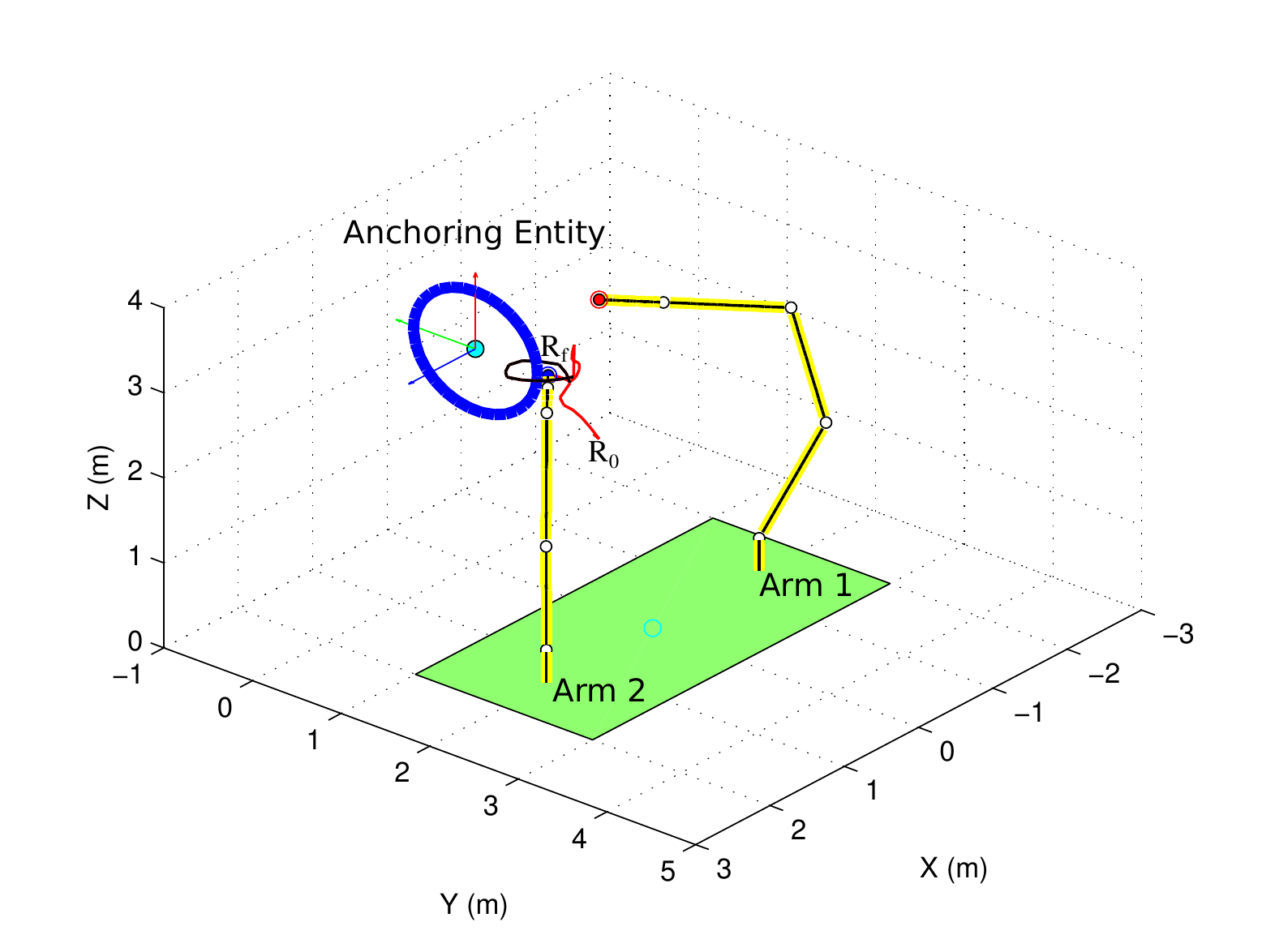}
                \vspace*{-0.7cm}
                \caption{Arm~2 Grasps $R_0$, Arm~1 Releases $R_0$}
                \vspace*{0.1cm}
                \label{fig:p8}
        \end{subfigure}%
        \begin{subfigure}[b]{0.33\textwidth}
                \includegraphics[width=\textwidth, trim={0cm 0cm 0cm 2cm}, clip]{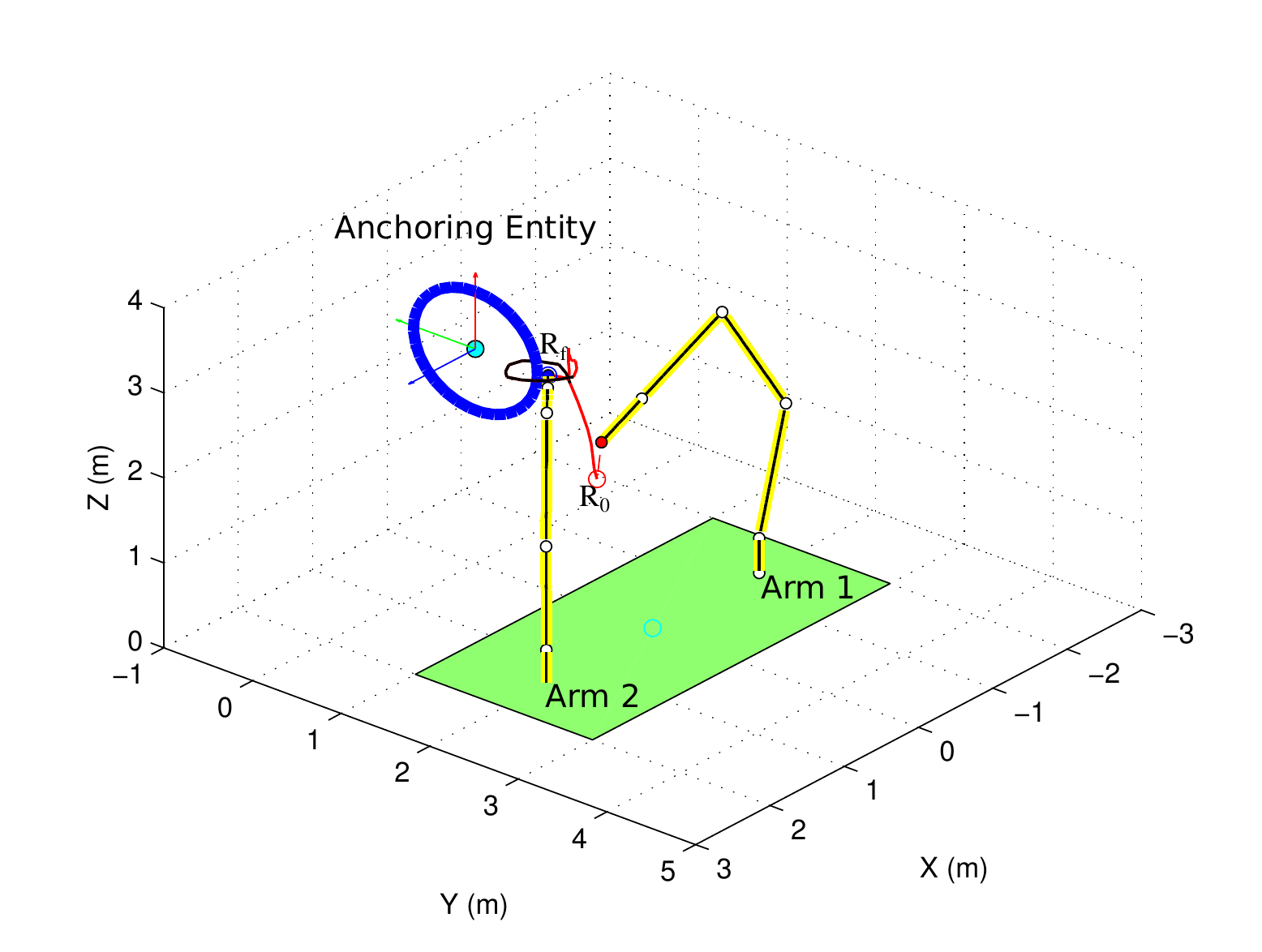}
                \vspace*{-0.7cm}
                \caption{Arm~1 Grasps $R_f$}
                \vspace*{0.1cm}
                \label{fig:p9}
        \end{subfigure}%
        \begin{subfigure}[b]{0.33\textwidth}
                \includegraphics[width=\textwidth, trim={0cm 0cm 0cm 2cm}, clip]{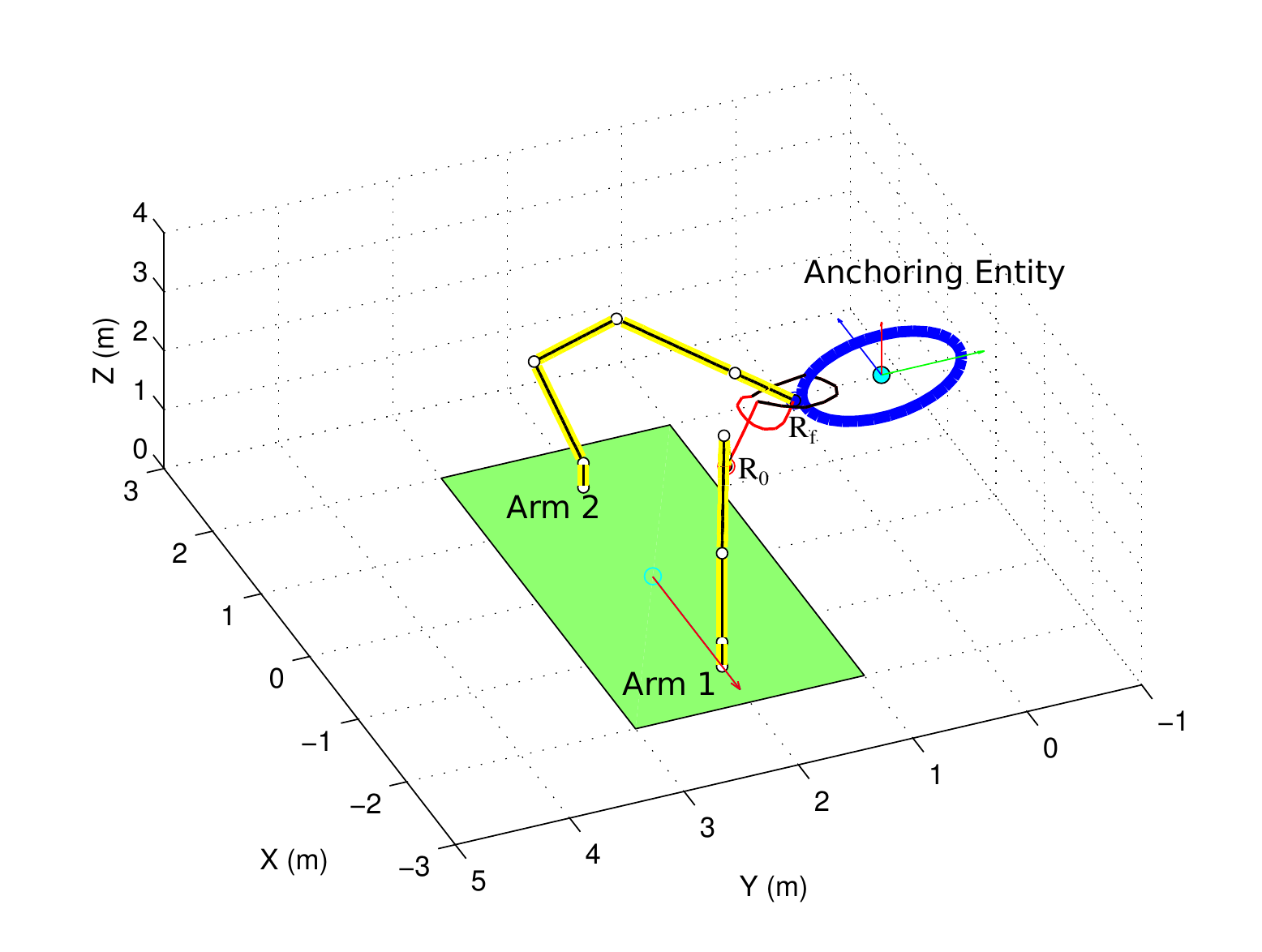}
                \vspace*{-0.7cm}
                \caption{Trefoil Loop Completed}
                \vspace*{0.1cm}
                \label{fig:p10}
        \end{subfigure}\\
        \caption{Step by step illustration of knotting a trefoil according to Sec.~\ref{ssec:trefoil_knot}.}
        \label{fig:knotting_all_steps}
\end{figure*}

\subsection{Beyond The Trefoil Knot}
\label{ssec:beyond}

The other knots in Fig.~\ref{fig:overview-table} can be created by adapting the action sequence for the trefoil knot from Sec.~\ref{ssec:trefoil_knot}. For the sake of brevity, below we omit which arm grasps which side of the rope and from which side the loop is approached. However, each step's action together with the final knot type yield an unequivocal assignment. The numbering of the steps below refers to the numbering of the steps in Section~\ref{sec:technique}:

\subsubsection{Unknot}
Executing steps {\footnotesize\emph{1, 2, 3, 4, 5}}

\subsubsection{Knot $3_1$}
Executing steps {\footnotesize\emph{1, 2, 3, 4, 5, 6, 7, 8, 9, 10}}

\subsubsection{Knot $4_1$}
Executing steps {\footnotesize\emph{1, 2, 3, 4, 5, \textbf{6}, \textbf{6}, 7, 8, 9, 10}}

\subsubsection{Knot $5_2$}
Executing steps {\footnotesize\emph{1, 2, 3, 4, 5, \textbf{6}, \textbf{6}, \textbf{6}, 7, 8, 9, 10}}

\subsubsection{Knot $7_3$}
Executing steps {\footnotesize\emph{1, \ldots, 5, 6, 6, 6, \textbf{7}, \textbf{8}, \textbf{9}, \textbf{7}, \textbf{8}, \textbf{9}, 10}}

In more detail:
The \emph{Unknot} is completed after passing the rope through the anchoring entity.
%
%
Knot $4_1$ requires the same steps as the trefoil but with two twists (i.e. step 6). 
Knot $5_2$ requires the same steps as the trefoil but with three twists (i.e. step 6). 
Knot $7_3$ requires the same steps as knot $5_2$, but with two insertions through the loop (i.e. steps 7, 8, 9). 

\subsection{Behavior Tree Formulation}
\label{ssec:bts}

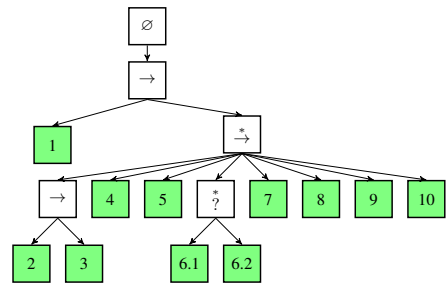
\begin{figure}[h]
  \begin{center}
\scalebox{0.7}{
  \begin{tikzpicture}[node distance=4cm,>=stealth',bend angle=20,auto,font=\small,style={align=center}]
    \tikzstyle{decision} = [diamond,draw,fill=blue!20,text badly centered,node distance=3cm,inner sep=0.5pt]

    \node (root)   [                             draw,thick,minimum width=0.7cm,minimum height=0.7cm] {$\varnothing$};
    \node (seq)    [          below=0.3cm of root,draw,thick,minimum width=0.7cm,minimum height=0.7cm] {$\rightarrow$};

    \node (act1)   [          below=0.50cm of seq,draw,thick,minimum width=0.7cm,minimum height=0.7cm, fill=green!50 , xshift=-1.8cm] {1};
    \node (seq1)    [         below=0.3cm of seq,draw,thick,minimum width=0.7cm,minimum height=0.7cm, xshift=+1.8cm] {$\overset{\ast}{\rightarrow}$};

    \node (seq11)   [         below=0.50cm of seq1,draw,thick,minimum width=0.7cm,minimum height=0.7cm,               xshift=-3.5cm] {$\rightarrow$};

    \node (act111)   [        below=0.50cm of seq11,draw,thick,minimum width=0.7cm,minimum height=0.7cm, fill=green!50 , xshift=-0.5cm] {2};
    \node (act222)   [        below=0.50cm of seq11,draw,thick,minimum width=0.7cm,minimum height=0.7cm, fill=green!50 , xshift=+0.5cm] {3};

    \node (act11)   [         below=0.50cm of seq1,draw,thick,minimum width=0.7cm,minimum height=0.7cm, fill=green!50 , xshift=-2.5cm] {4};

    \node (act22)   [         below=0.50cm of seq1,draw,thick,minimum width=0.7cm,minimum height=0.7cm, fill=green!50 , xshift=-1.5cm] {5};

    \node (sel11)   [         below=0.50cm of seq1,draw,thick,minimum width=0.7cm,minimum height=0.7cm,               xshift=-0.5cm] {$\overset{\ast}{?}$};
    
    \node (act333)   [        below=0.50cm of sel11,draw,thick,minimum width=0.7cm,minimum height=0.7cm, fill=green!50 , xshift=-0.5cm] {6.1};
    
    \node (act444)   [        below=0.50cm of sel11,draw,thick,minimum width=0.7cm,minimum height=0.7cm, fill=green!50 , xshift=+0.5cm] {6.2};

    \node (act33)   [         below=0.50cm of seq1,draw,thick,minimum width=0.7cm,minimum height=0.7cm, fill=green!50 , xshift=+0.5cm] {7};

    \node (act44)   [         below=0.50cm of seq1,draw,thick,minimum width=0.7cm,minimum height=0.7cm, fill=green!50 , xshift=+1.5cm] {8};

    \node (act55)   [         below=0.50cm of seq1,draw,thick,minimum width=0.7cm,minimum height=0.7cm, fill=green!50 , xshift=+2.5cm] {9};

    \node (act66)   [         below=0.50cm of seq1,draw,thick,minimum width=0.7cm,minimum height=0.7cm, fill=green!50 , xshift=+3.5cm] {10};

	\draw [->] (root.south)  -- (seq.north);
	\draw [->] (seq.south)  -- (act1.north);
	\draw [->] (seq.south)  -- (seq1.north);
	\draw [->] (seq1.south)  -- (seq11.north);
	\draw [->] (seq1.south)  -- (act11.north);
	\draw [->] (seq1.south)  -- (act22.north);
	\draw [->] (seq1.south)  -- (sel11.north);
	\draw [->] (seq1.south)  -- (act33.north);
	\draw [->] (seq1.south)  -- (act44.north);
	\draw [->] (seq1.south)  -- (act55.north);
	\draw [->] (seq1.south)  -- (act66.north);
	\draw [->] (seq11.south)  -- (act111.north);
	\draw [->] (seq11.south)  -- (act222.north);
	\draw [->] (sel11.south)  -- (act333.north);
	\draw [->] (sel11.south)  -- (act444.north);

  \end{tikzpicture}
}
\end{center}
  \caption{BT representation of the knotting task for the trefoil knot. Numbered boxes represent knotting actions within their context; for reference see the steps in Sec.~\ref{ssec:trefoil_knot}.}
  \label{fig:bt_trefoil}
\end{figure}

The hybrid dynamics described in Sec.~\ref{ssec:trefoil_knot} and Sec.~\ref{ssec:beyond} can be conveniently instantiated in the form of a Behavior Tree (BT)~\cite{ale_icra2013}. Even though BTs were not used to implement the simulations in Fig.~\ref{fig:knotting_all_steps} or the experiments with the NAO in Sec.~\ref{sec:nao}, the hierarchical structure of BTs provides a convenient representation for the knotting task, see Fig.~\ref{fig:bt_trefoil}:
\emph{i)} step 1 must move the base to track the anchoring entity's motion throughout the whole task, represented using a so-called \emph{Sequence node}; 
\emph{ii)} steps 6.1 and 6.2 are alternative ways of achieving the same goal, represented using a so-called \emph{Selector* node}; 
\emph{iii)} step 3 could fall back to re-grasping the rope using step 2 if it was dropped during the insertion step, represented by another \emph{Sequence node}.

\section{Experimental Evaluation}
\label{sec:results}

To test the viability of our approach to rope insertion and knotting, we first analyse the robustness of the insertion action disregarding robot and rope dynamics (Sec.~\ref{sec:results-isolation}), then we test the controller of the whole robot with the rope in a Matlab knotting simulation (Sec.~\ref{sec:results-simulation}), lastly we implement the knotting system on a NAO robot. Illustrative real-time videos of the experiments described here are found under 
\url{http://www.csc.kth.se/~almc/humanoids16/}.

\subsection{Insertion Action in Isolation}
\label{sec:results-isolation}

To perform the insertion action, the robot must move the manipulator which holds the rope into the loop, stop, and re-grasp the rope with another manipulator. Below, we study how our insertion action copes with different shapes of the target loop.

\subsubsection{Stopping}

The switching condition for the insertion action needs to stop the manipulator when its trajectory has passed through the loop. In Sec.~\ref{sec:insertion-control} we identify this condition by a drop in magnetic field intensity. We validate this formulation by simulating the trajectory of a manipulator for static target loops of different (non-trivial) shapes. We consider a planar loop (for reference), a 90$^\circ$ folded planar loop, and a double loop, as sketched in Fig.~\ref{fig:overview}. 

Results are displayed in Fig.~\ref{fig:stopping}, where black lines indicate the manipulator trajectories. In all cases the trajectories stop centrally in the loop, confirming the switching condition.



\subsubsection{Deformation}

During execution of an insertion action with a real rope, the target loop might flex, bend, and change its size. Further, its position is subject to noise in perception. For this reason, we investigate the performance of our insertion action when the target loop deforms randomly during execution. We consider a \emph{planar} loop and compare  \emph{insertion quality} defined as the distance between the insertion point and the center of the loop and \emph{insertion delay} measured as the time it took (in number of iterations) for the particle to reach the plane of the loop. Further, we trace the trajectory and register whether it passed inside the loop (\emph{successful insertion}).

\begin{figure*}[]
  \centering
  \begin{subfigure}[b]{0.33\linewidth}
	\includegraphics[width=\textwidth]{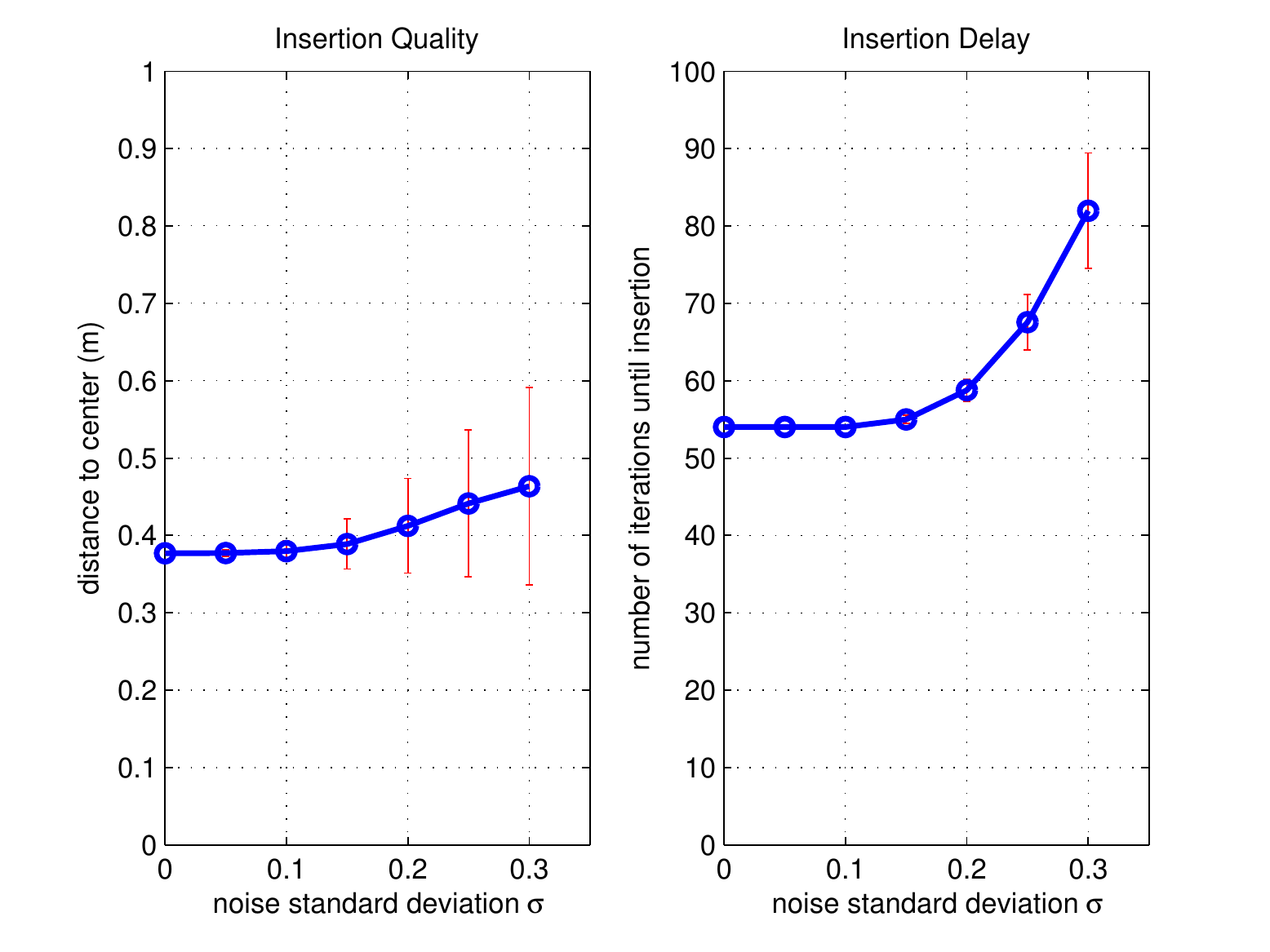}
    \caption{$\alpha = 1, \beta = 1$.}
    \label{fig:a1b1_isotropic}
  \end{subfigure}%
  \begin{subfigure}[b]{0.33\linewidth}
    \includegraphics[width=\textwidth]{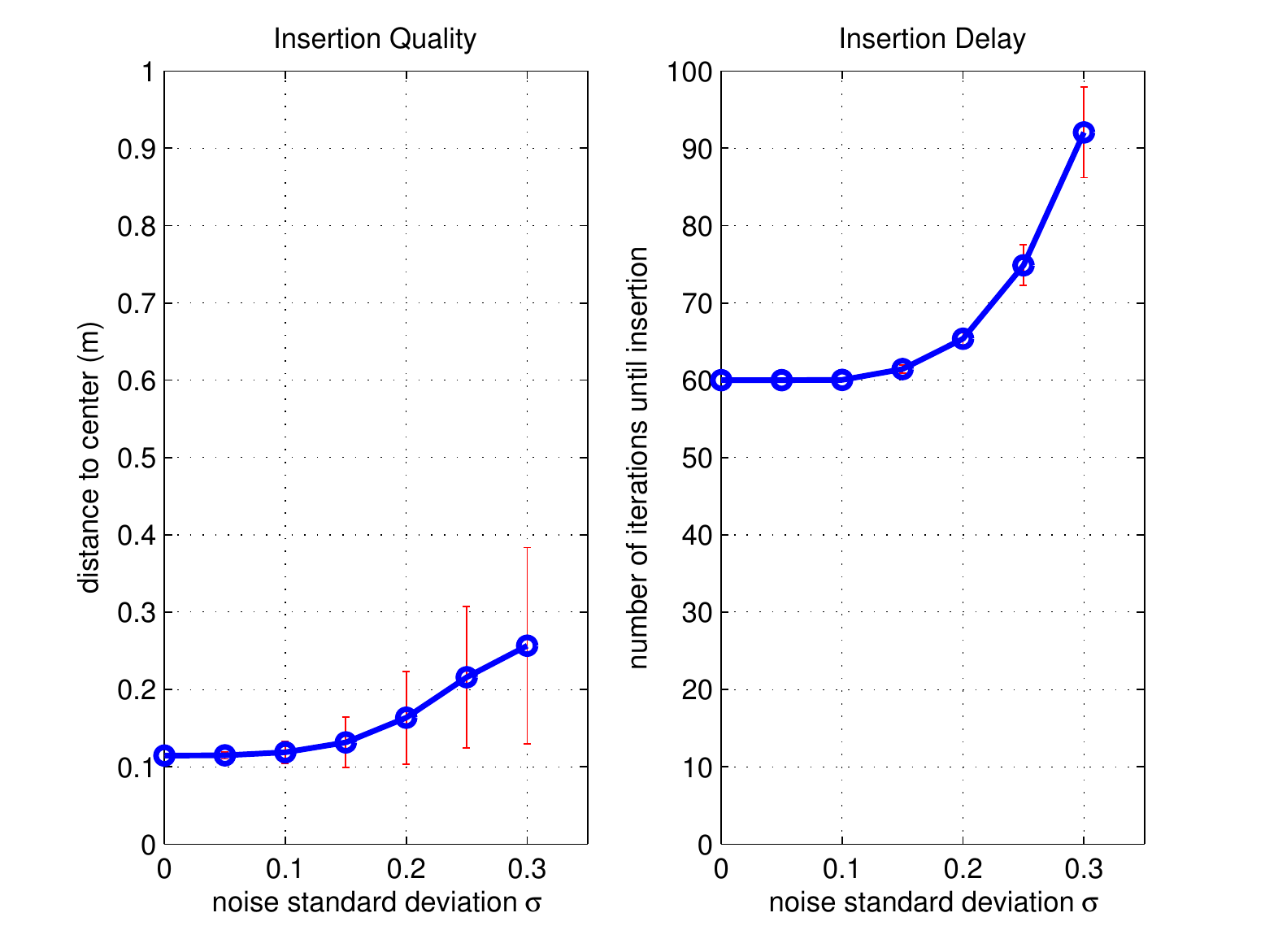}
    \caption{$\alpha = 2, \beta = 1$.}
    \label{fig:a2b1_isotropic}
    \end{subfigure}%
  \begin{subfigure}[b]{0.33\linewidth}
    \includegraphics[width=\textwidth]{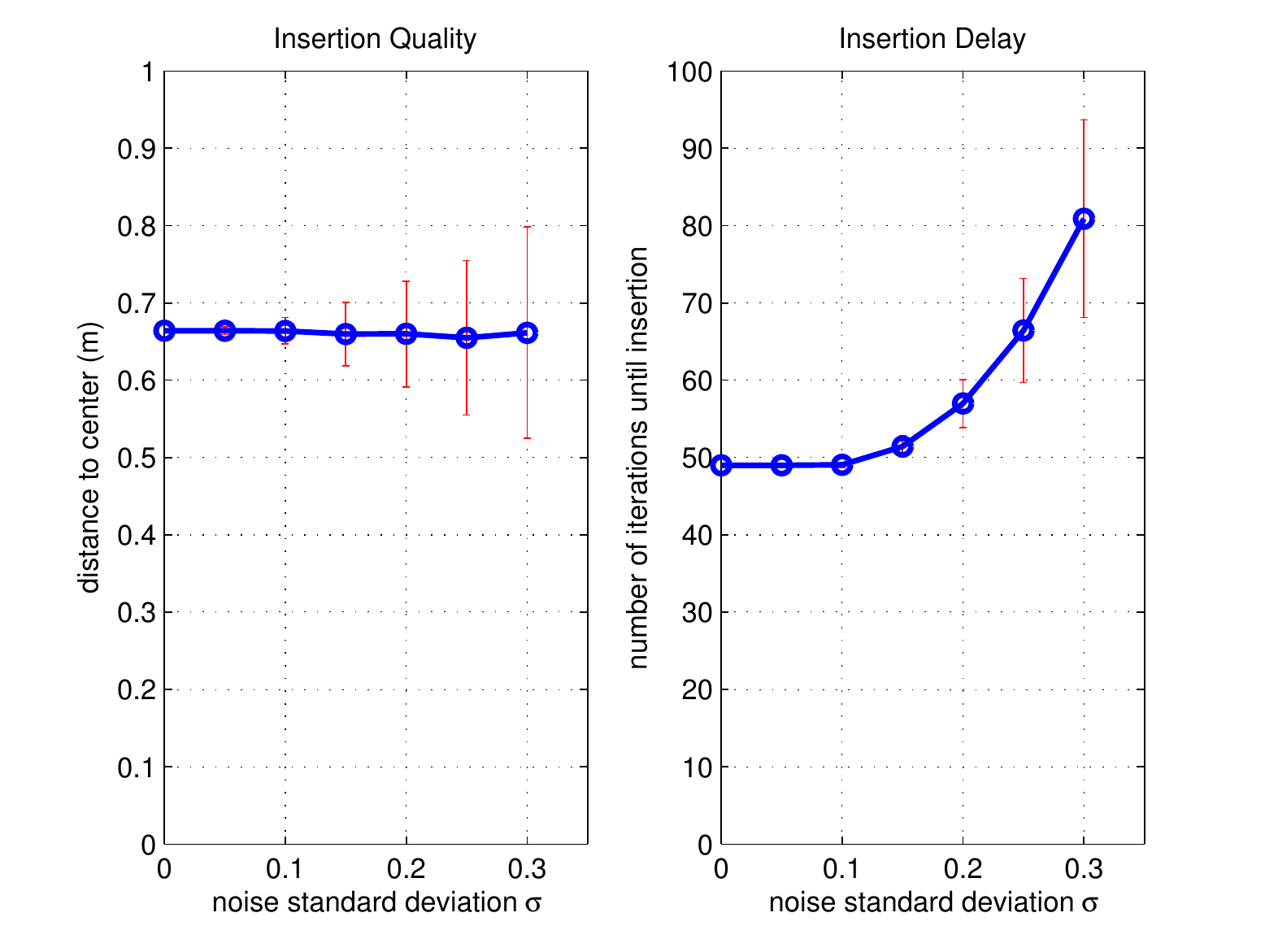}
    \caption{$\alpha = 1, \beta = 2$.}
    \label{fig:a1b2_isotropic}
    \end{subfigure}
  \caption{\emph{Insertion Quality} and \emph{Insertion Delay} plots for isotropic noise using three different $\alpha,\beta$ combinations.}
  \label{fig:alpha_beta_isotropic}
\end{figure*}

\begin{figure*}[]
  \centering
  \begin{subfigure}[b]{0.33\linewidth}
	\includegraphics[width=\textwidth]{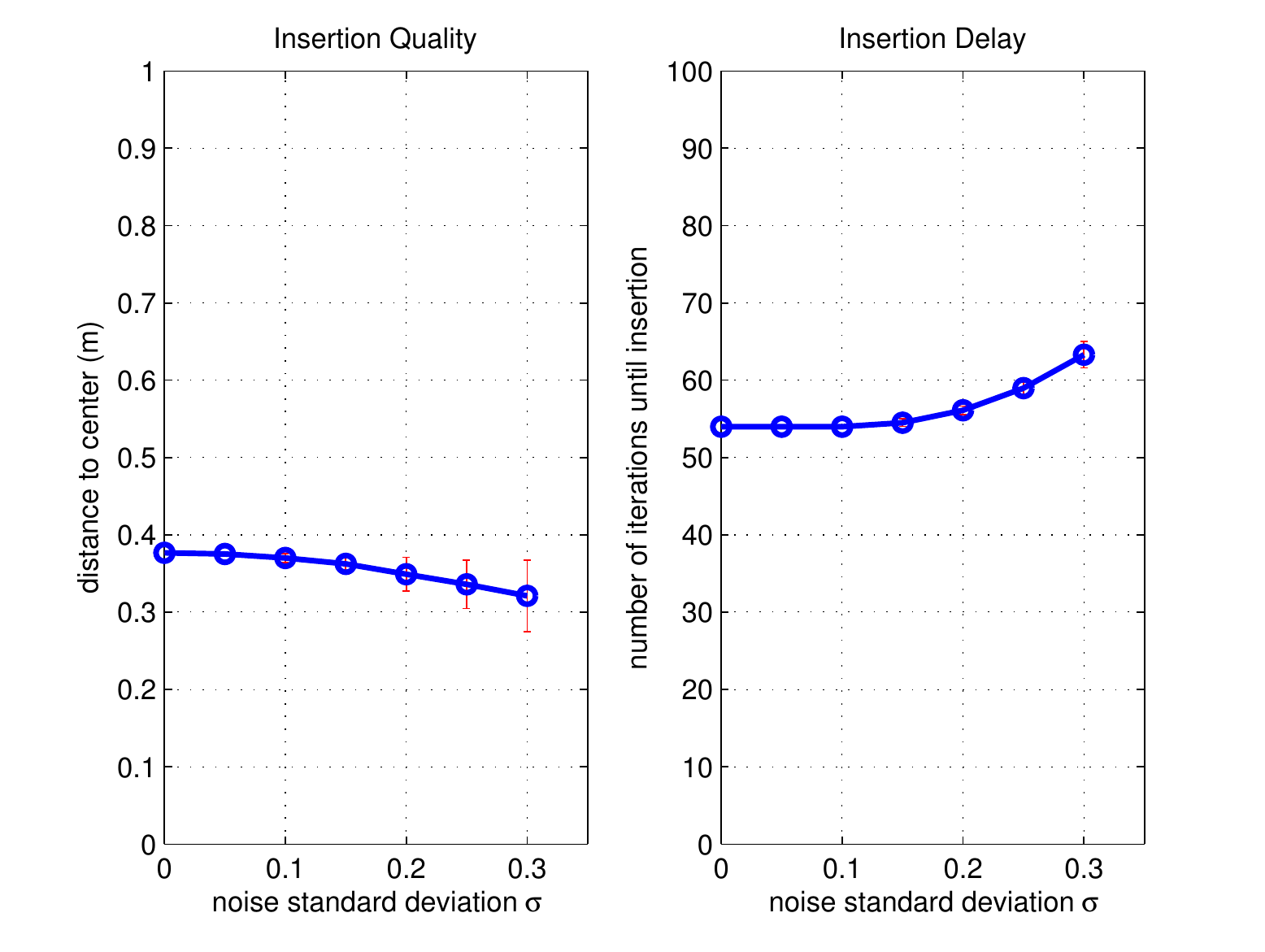}
    \caption{$\alpha = 1, \beta = 1$.}
    \label{fig:a1b1_radial}
  \end{subfigure}%
  \begin{subfigure}[b]{0.33\linewidth}
    \includegraphics[width=\textwidth]{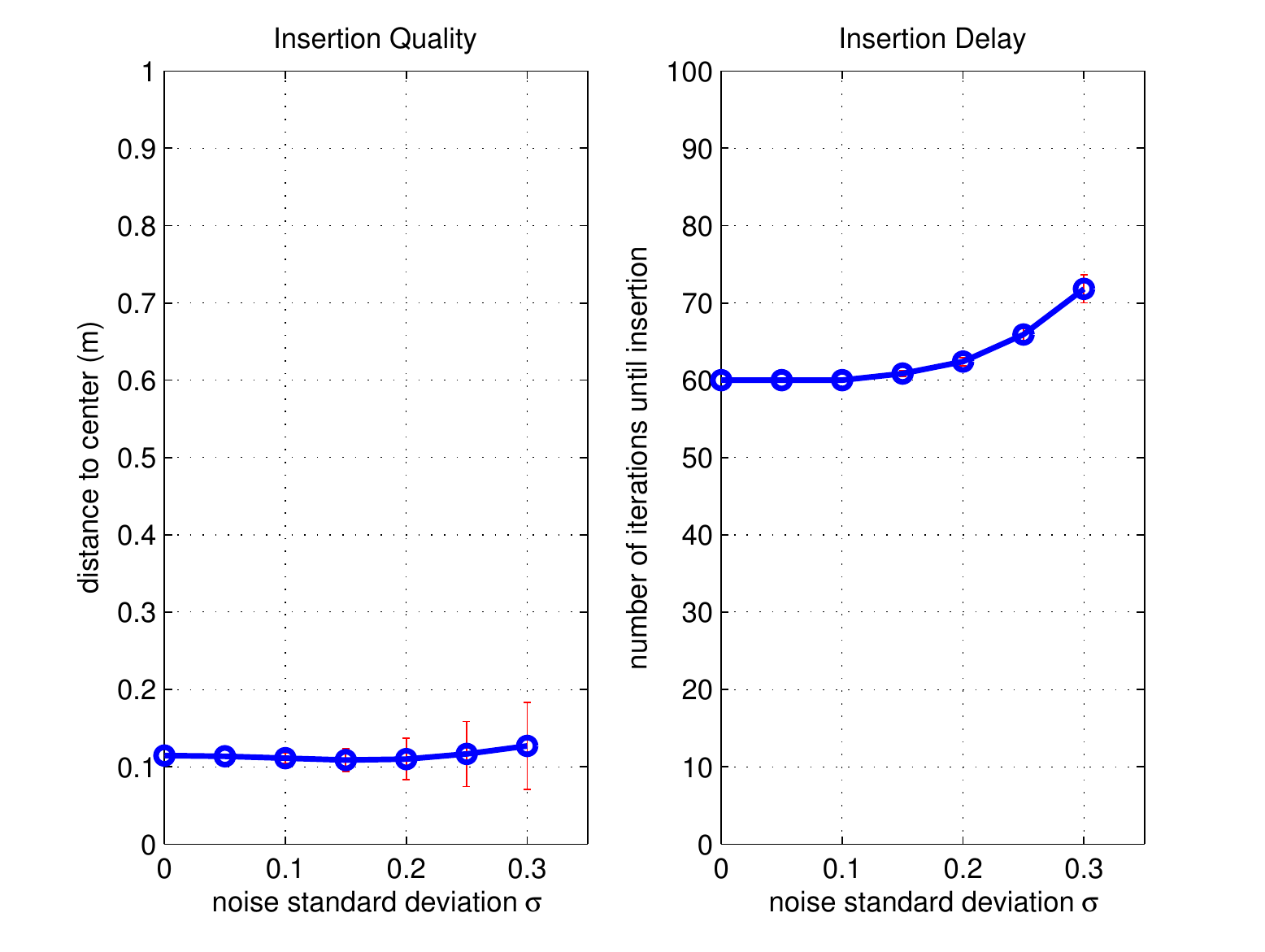}
    \caption{$\alpha = 2, \beta = 1$.}
    \label{fig:a2b1_radial}
    \end{subfigure}%
  \begin{subfigure}[b]{0.33\linewidth}
    \includegraphics[width=\textwidth]{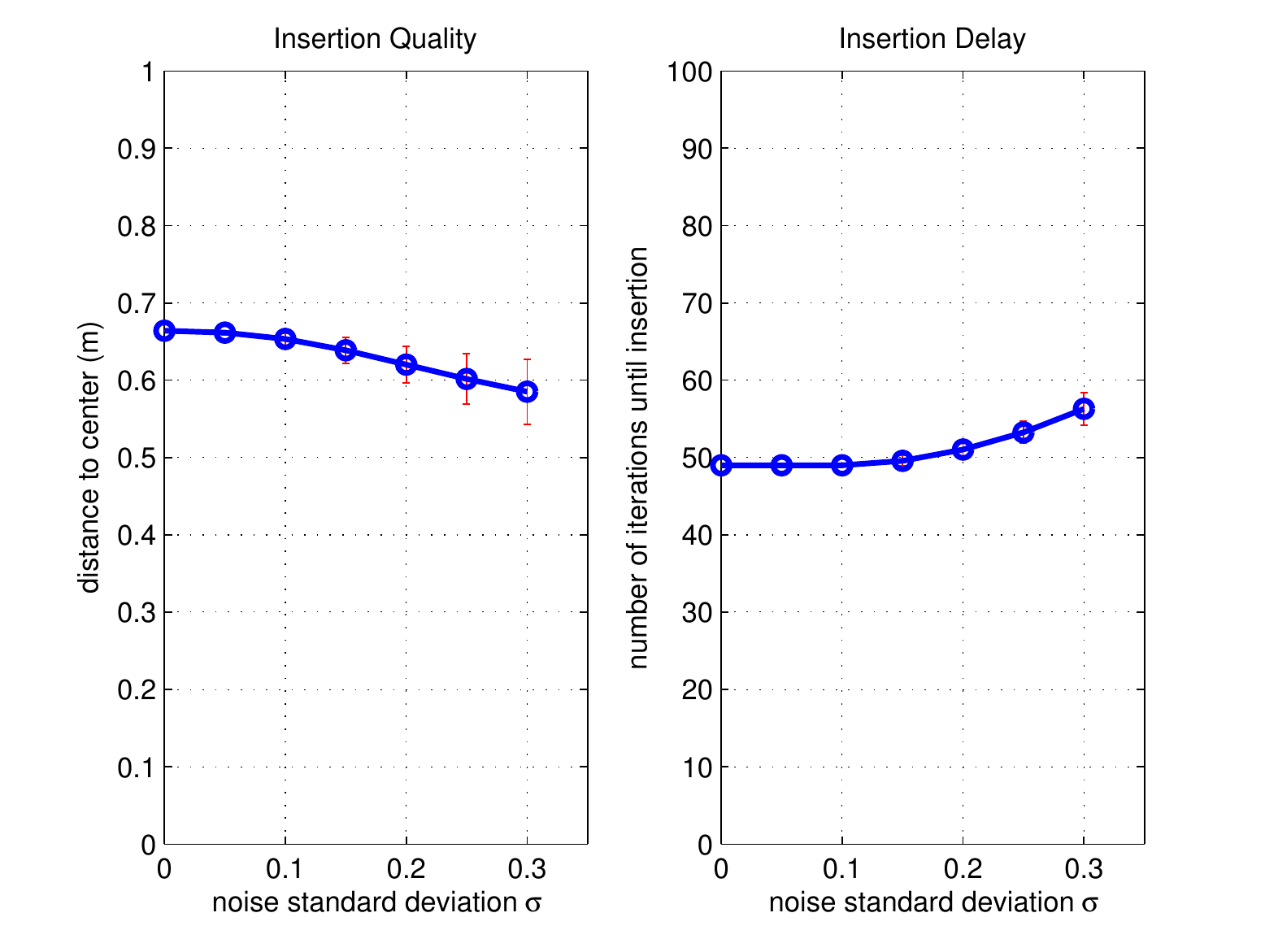}
    \caption{$\alpha = 1, \beta = 2$.}
    \label{fig:a1b2_radial}
    \end{subfigure}
  \caption{\emph{Insertion Quality} and \emph{Insertion Delay} plots for radial noise using three different $\alpha,\beta$ combinations.}
  \label{fig:alpha_beta_radial}
\end{figure*}

For the experiments, we start with a circular planar loop of radius 1~m (discretized to $0.1$ radians) laying in the $X,Y$-plane and perturb vertex positions in each time step. The perturbation consists of (a) isotropic Gaussian noise and (b) cylindrical Gaussian noise for the radial direction and the $Z$-axis with standard deviation between $0.0$ and $0.3$ in steps of $0.05$. We record 1000 trajectories (for each noise setting) from the same initial location and plot noise intensity against \emph{insertion quality} and the \emph{insertion delay} in Fig.~\ref{fig:alpha_beta_isotropic} and Fig.~\ref{fig:alpha_beta_radial} for different values of $\alpha$ and $\beta$.


The figures show that \emph{insertion delay} increases with higher intensities of noise for both isotropic and radial perturbation. \emph{Insertion quality} decreases with higher intensities of noise (greater distances) for isotropic noise. However, \emph{insertion quality} increases with higher levels of cylindrical noise.
Increasing $\alpha$, consistently yields insertions with better quality but taking slightly longer to complete. Conversely, increasing $\beta$, consistently decreases \emph{insertion quality} but finishes slightly faster. 
Most relevant, for all the 42.000 insertion attempts, we observed only 14 failed insertions (occurring at the highest noise setting). 

%
%
%

\subsection{Insertion Action in Context}
\label{sec:results-simulation}

In reality, our approach needs to control a complete robot that manipulates a flexible body of rope. Further, both rope and anchoring entity can move and deform during execution of an insertion action. This makes it relevant to validate our insertion action in presence of rope and robot dynamics. Instead of a single particle as in Sec.~\ref{sec:results-isolation}, in this experiment, we simulate a length of rope and a two-armed robot with a mobile base and create trefoil knots. For more information about the simulator head to: \url{www.github.com/~almc}

\subsubsection{Deformation of The Loop}

\begin{figure}[h]
  \centering
  \begin{subfigure}[b]{0.4\linewidth}
	\includegraphics[width=\textwidth, trim={6cm 3.5cm 3cm 2cm},clip]{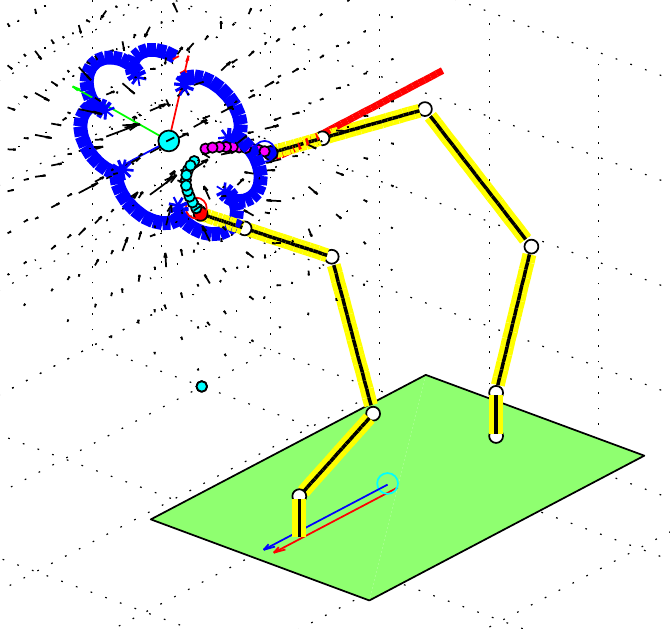}
    \caption{Parallel deformation.}
    \label{fig:def1}
  \end{subfigure}%
  \begin{subfigure}[b]{0.4\linewidth}
    \includegraphics[width=\textwidth,trim={6cm 3.5cm 3cm 2cm},clip]{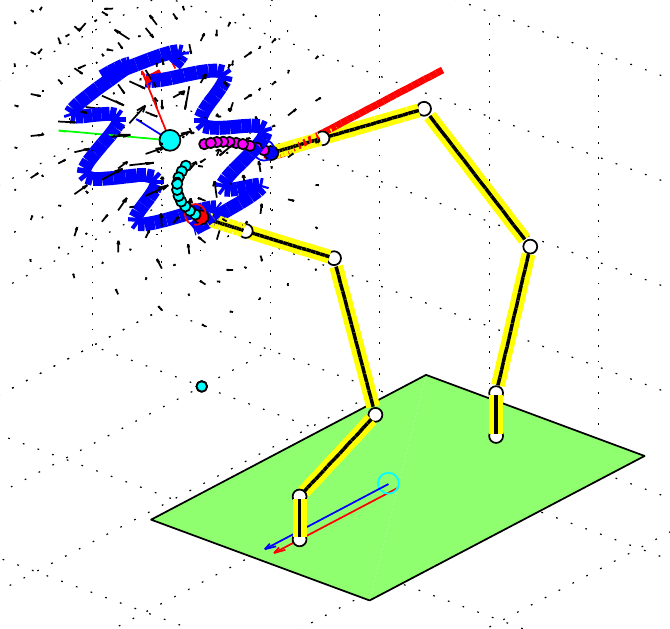}
    \caption{Perpendicular deformation}
    \label{fig:def2}
    \end{subfigure}
  \caption{Anchoring entity deformation: parallel and perpendicular to loop.}
  \label{fig:deformation}
\end{figure}

In this experiment, we deform the anchoring entity's loop by adding a cosine wave of time varying frequency to its coordinates (either parallel or perpendicular to the loop's plane) as seen in Fig.~\ref{fig:deformation}.

We observe that the wave's frequency does not have an effect on the success or failure of the knot. However, extremely high frequencies require a more fine grained discretization of the anchoring entity which in turn increases the controller's computation time. In detail, we observe that the ratio between the amplitude of the wave and the loop's radius, $\mathcal{R}^w_l$, is a limiting factor for the controller. Setting the amplitude to $1/5$ of the loop's radius always yields successful knots. Increasing the deformation beyond $\mathcal{R}^w_l > 1/5$ is also successful up to $\mathcal{R}^w_l = 1/2$ but requires a more fine-grained model of the loop in our simulation.

\subsubsection{Translation \& Rotation of The Loop}

In this experiment, we move the loop at various velocities following different trajectories. Regardless of the trajectory described by the loop, we observe that the knotting is successful as long as the velocity and acceleration are bounded. The threshold depends on the parameters of the base controllers and is robot dependent. This shows qualitatively that our approach can knot with loops that \emph{i)} do not move too fast for the robot and controller, and \emph{ii)} do not change their velocity too abruptly.



\subsection{Knotting with a NAO Robot}
\label{sec:nao}

\begin{figure}[h]
\centering
\includegraphics[height = 0.4 \linewidth]{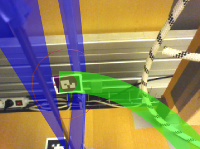}
\includegraphics[height = 0.4 \linewidth]{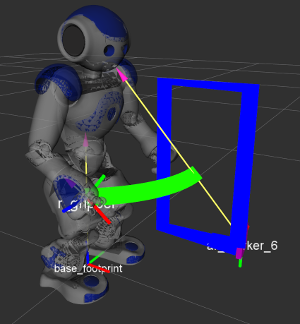}
\caption{\emph{Left:} The robot's camera perspective with visual markers, the detected anchoring entity (blue) and the manipulator trajectory (green). \emph{Right:} Same scene showing robot pose, detected loop (anchoring entity), and manipulator trajectory (computed following the magnetic field).}
\label{fig:nao-camera}
\end{figure}

In this part we want to find out whether our insertion action and knotting system are feasible on a humanoid robot platform. To this end, we conduct experiments with a NAO humanoid robot which additionally poses the problem of perceiving and following the rope and anchoring entity. For the experiments below, we provide the robot with visual markers to simplify the perception problem (see Figs.~\ref{fig:nao-camera},~\ref{fig:nao-unknot} and~\ref{fig:nao-trefoil}) and use visual servoing to move the robot's arms. Because of the NAO's limited manipulation abilities, in the trefoil case we manually hold the rope crossing point as seen in Fig.~\ref{fig:nao-trefoil}.

\subsubsection{Dynamic Loop}
In this experiment, we test the dynamic adaptability of the magnetic field to cope with deformations in the target loop. We disregard temporarily the knotting task to focus only on insertion. We placed 5 markers around the target loop and instructed the NAO to insert through it.

We observed that the trajectory of the arm would continuously adapt to account for changes in the deformation of the loop as seen in Fig.~\ref{fig:nao-dynamic}. The limitations in the speed of adaptation came from the software infrastructure used (ROS topic communication) and not from the computation of the magnetic field itself. For a live demonstration of this experiment refer to the accompanying videos.

\subsubsection{Knotting The Unknot and The Trefoil}

\begin{figure*}
\centering

\includegraphics[width = 0.15 \linewidth, trim={17cm 5cm 23cm 5cm}, clip]{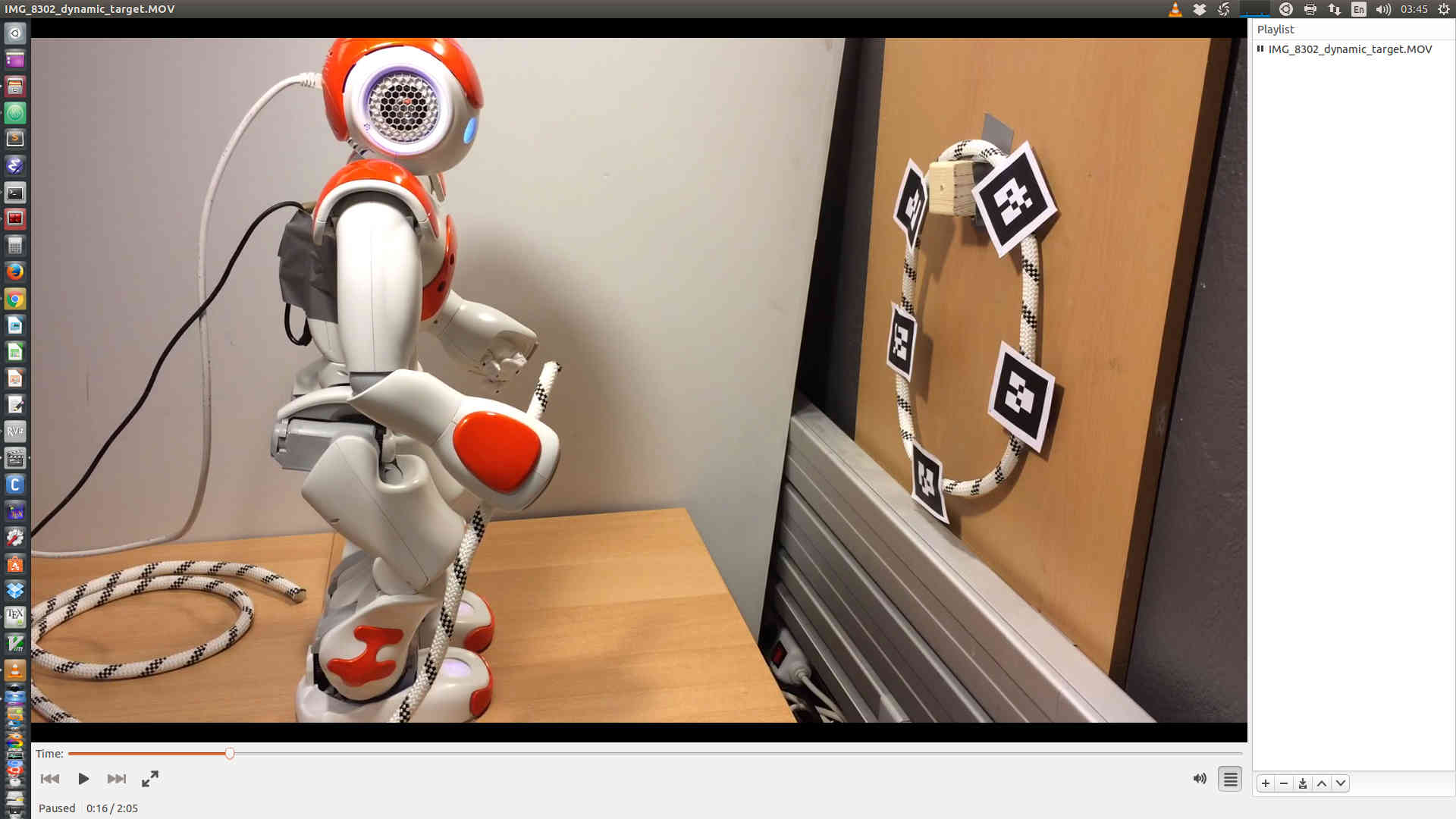}
\includegraphics[width = 0.15 \linewidth, trim={30cm 5cm 10cm 5cm}, clip]{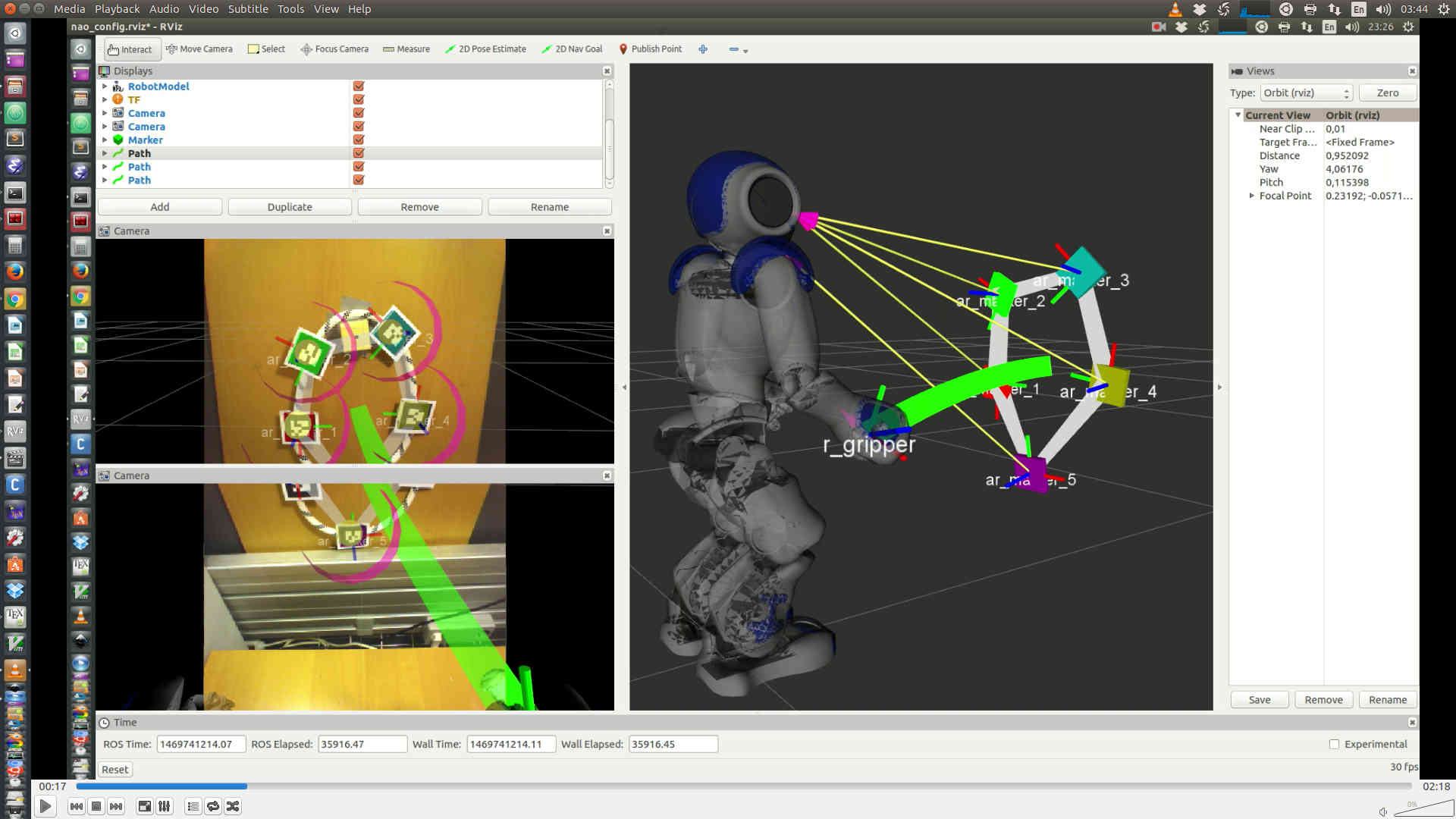}
\includegraphics[width = 0.15 \linewidth, trim={15cm 5cm 25cm 5cm}, clip]{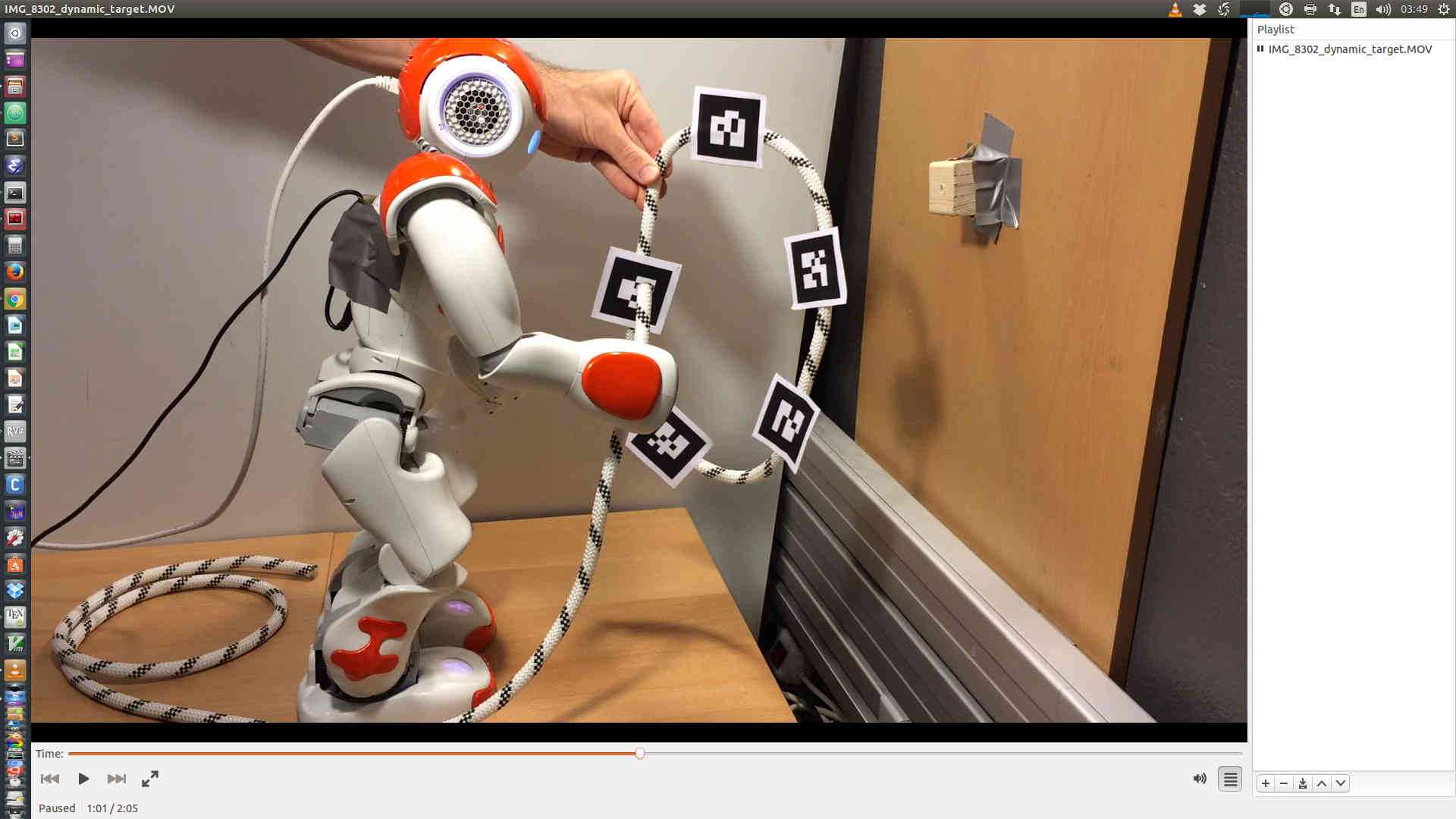}
\includegraphics[width = 0.15 \linewidth, trim={30cm 5cm 10cm 5cm}, clip]{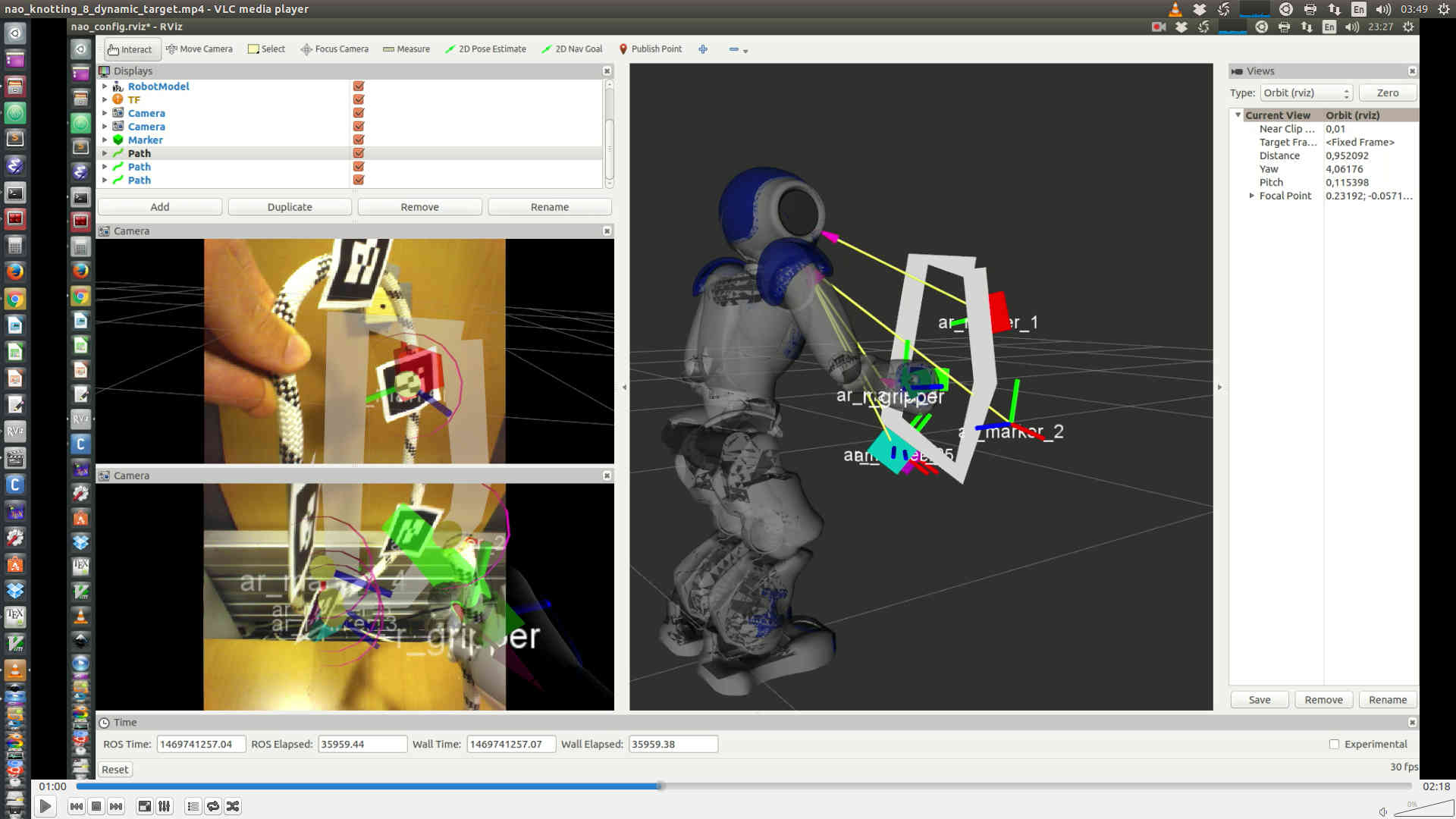}
\includegraphics[width = 0.15 \linewidth, trim={20cm 5cm 20cm 5cm}, clip]{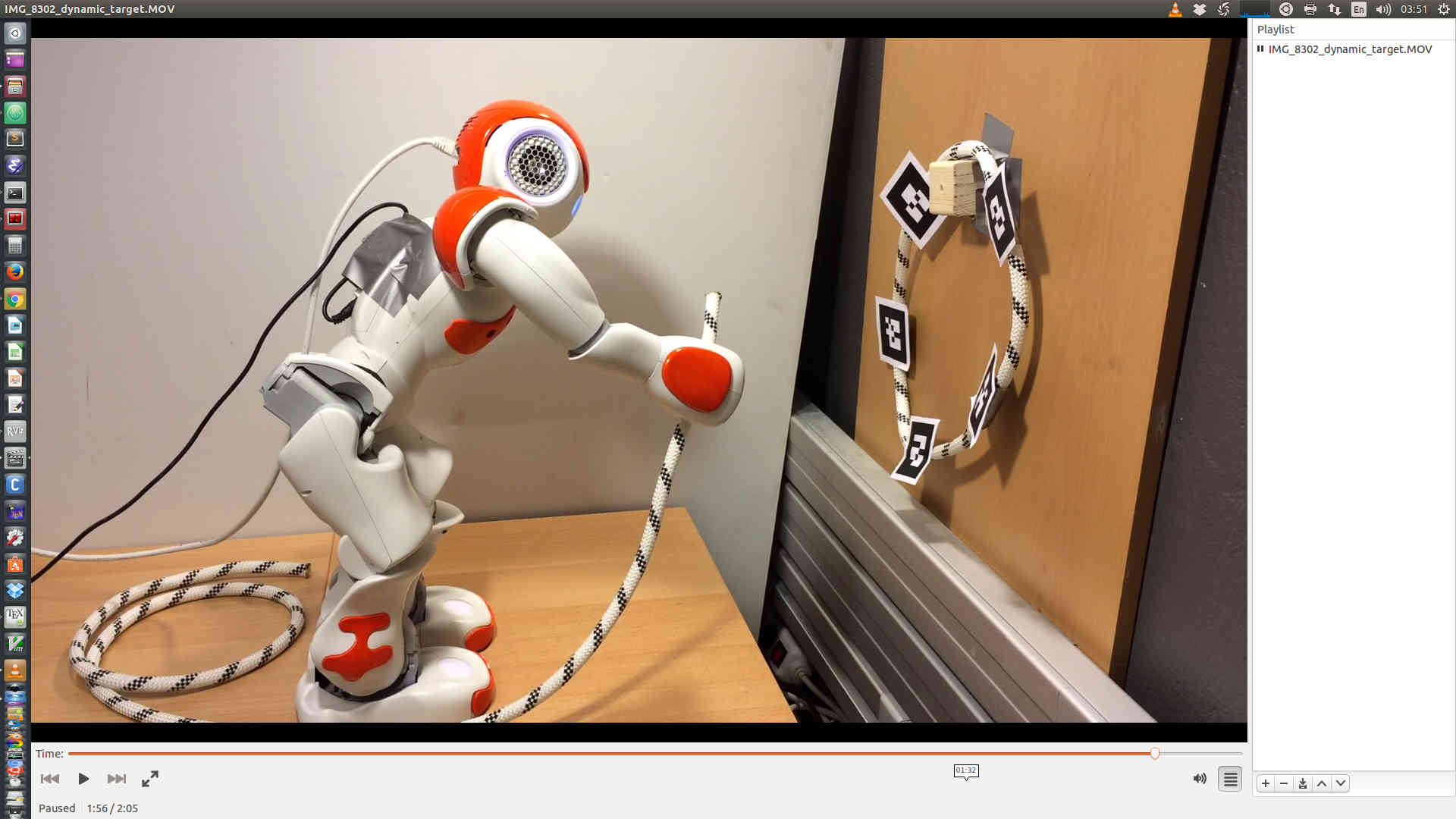}
\includegraphics[width = 0.15 \linewidth, trim={30cm 5cm 10cm 5cm}, clip]{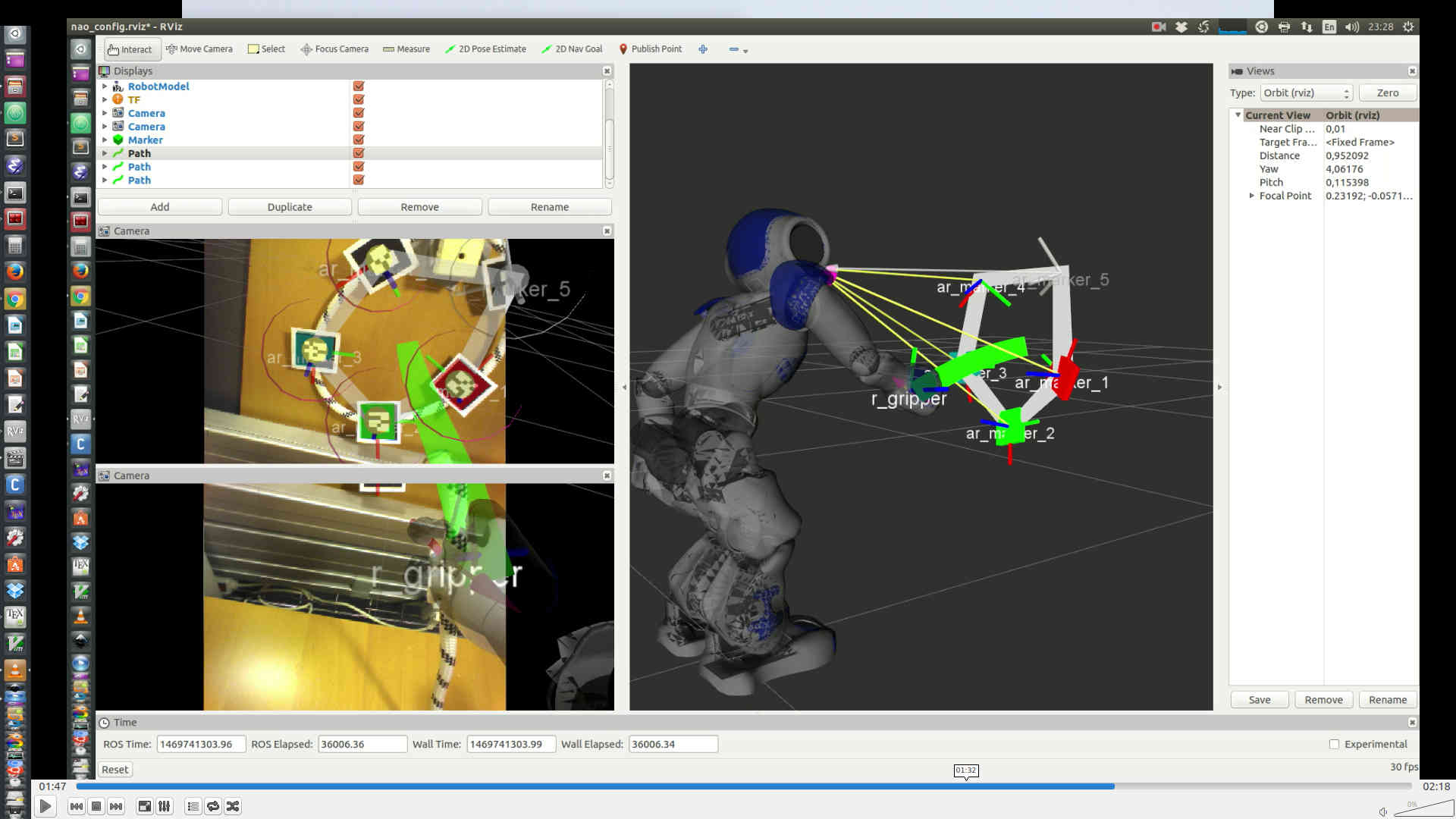}
\caption{NAO humanoid robot inserting through a dynamically deforming loop. For this, the NAO has to localize the target loop and recompute the magnetic field direction on each iteration of the controller. Each pair of figures from left to right corresponds to the same time instant.}
\label{fig:nao-dynamic}

\includegraphics[width = 0.15 \linewidth, trim={22cm 0cm 10cm 0cm}, clip]{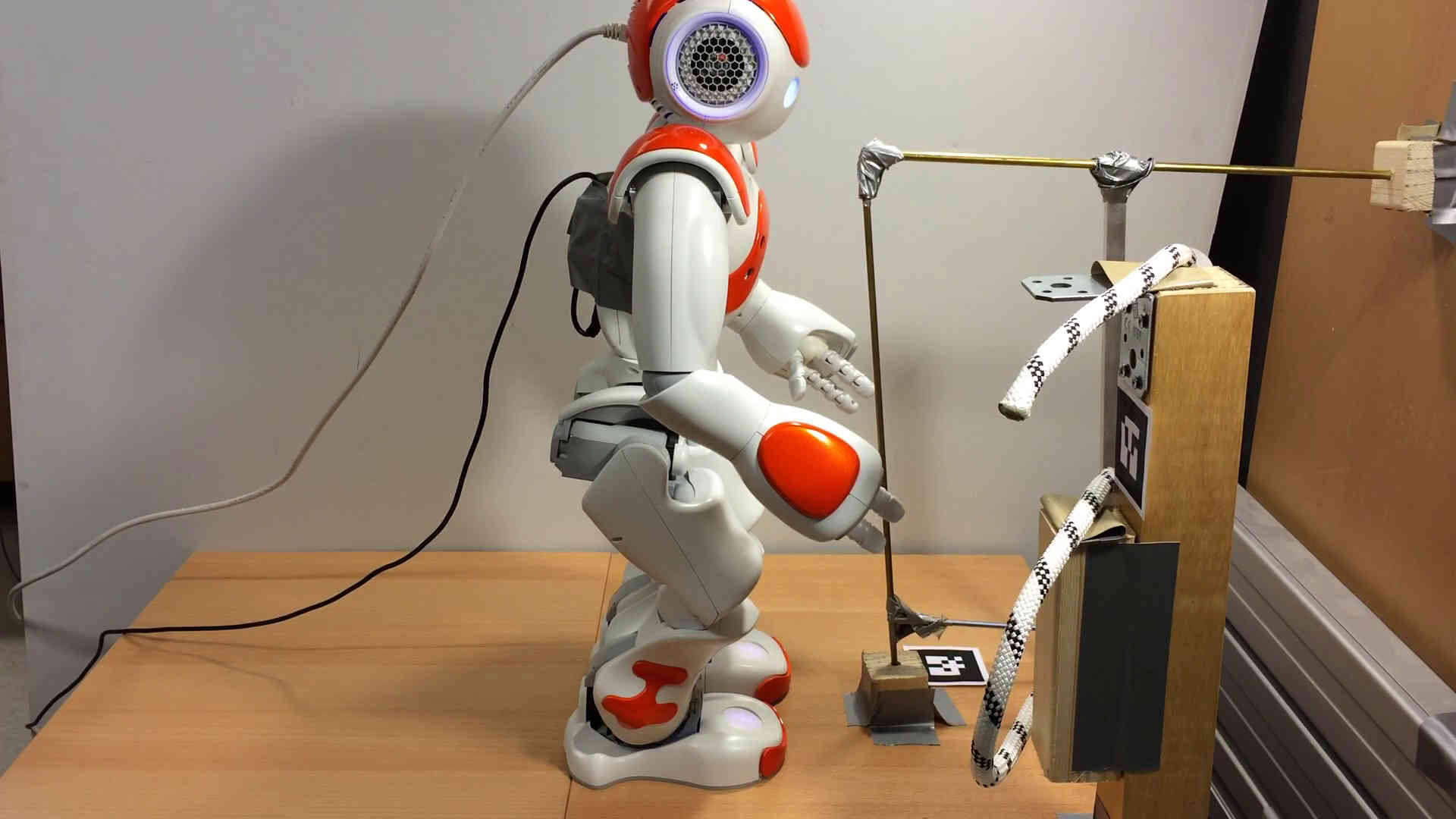}
\includegraphics[width = 0.15 \linewidth, trim={22cm 0cm 10cm 0cm}, clip]{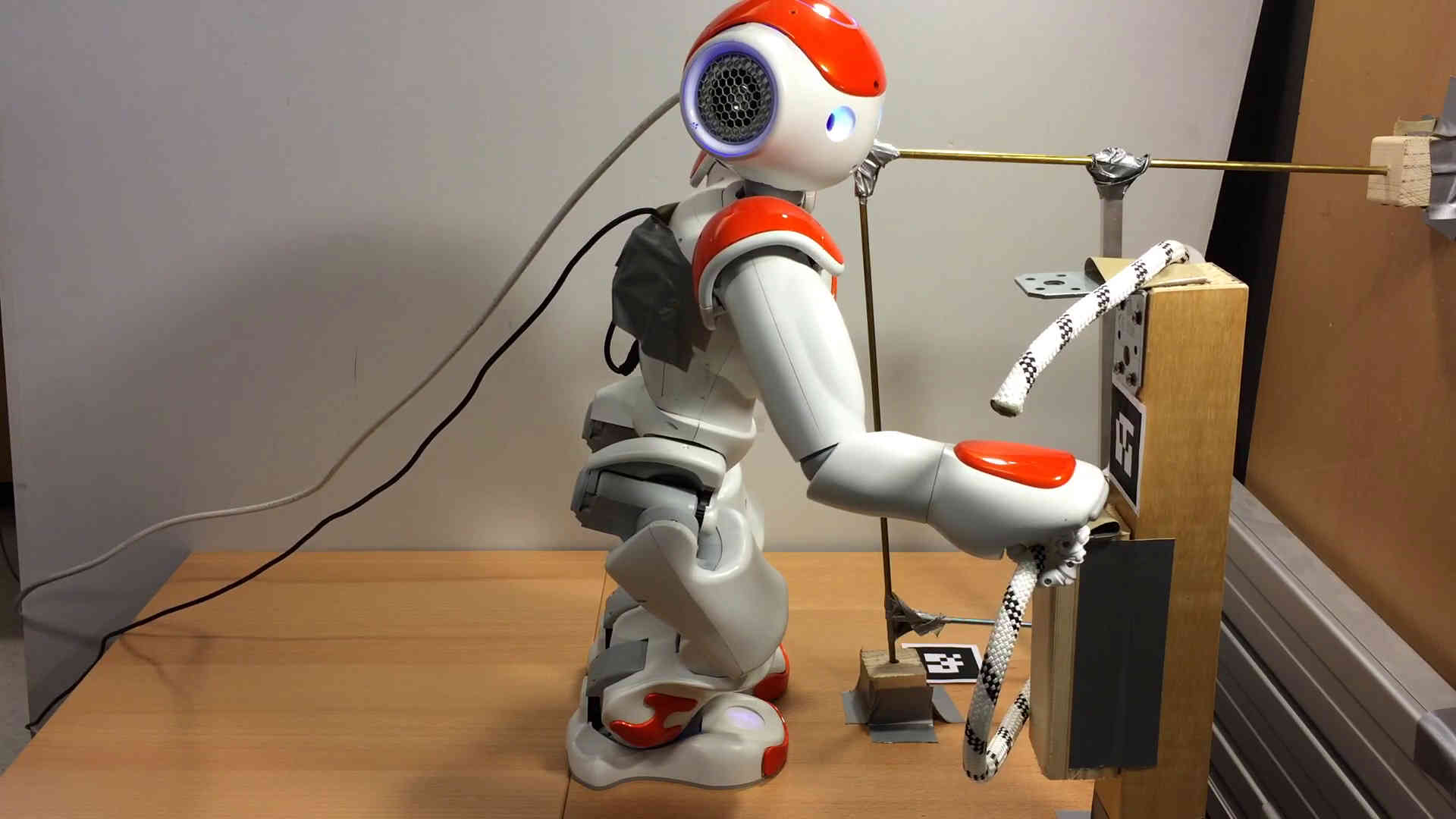}
\includegraphics[width = 0.15 \linewidth, trim={22cm 0cm 10cm 0cm}, clip]{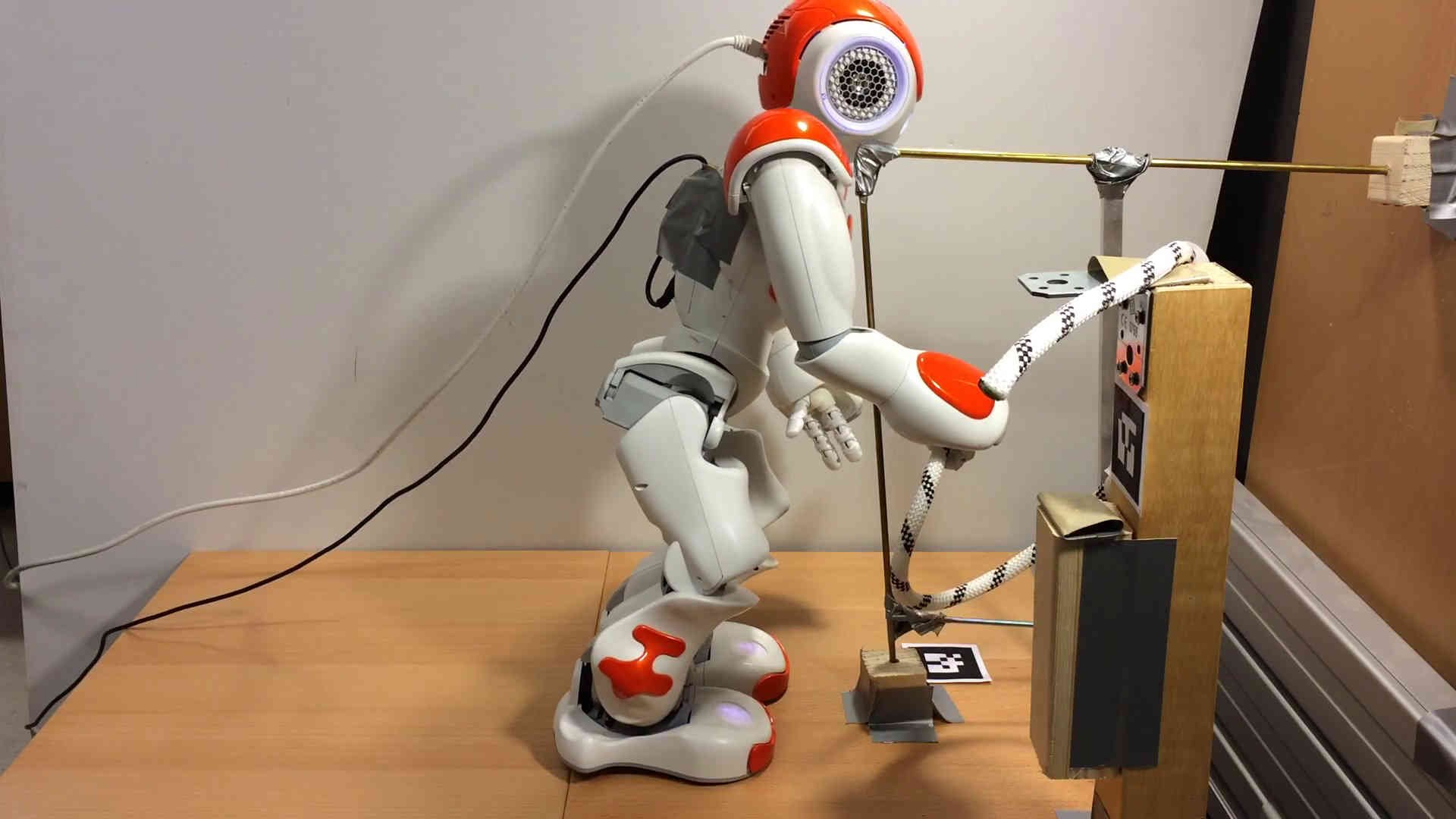}
\includegraphics[width = 0.15 \linewidth, trim={22cm 0cm 10cm 0cm}, clip]{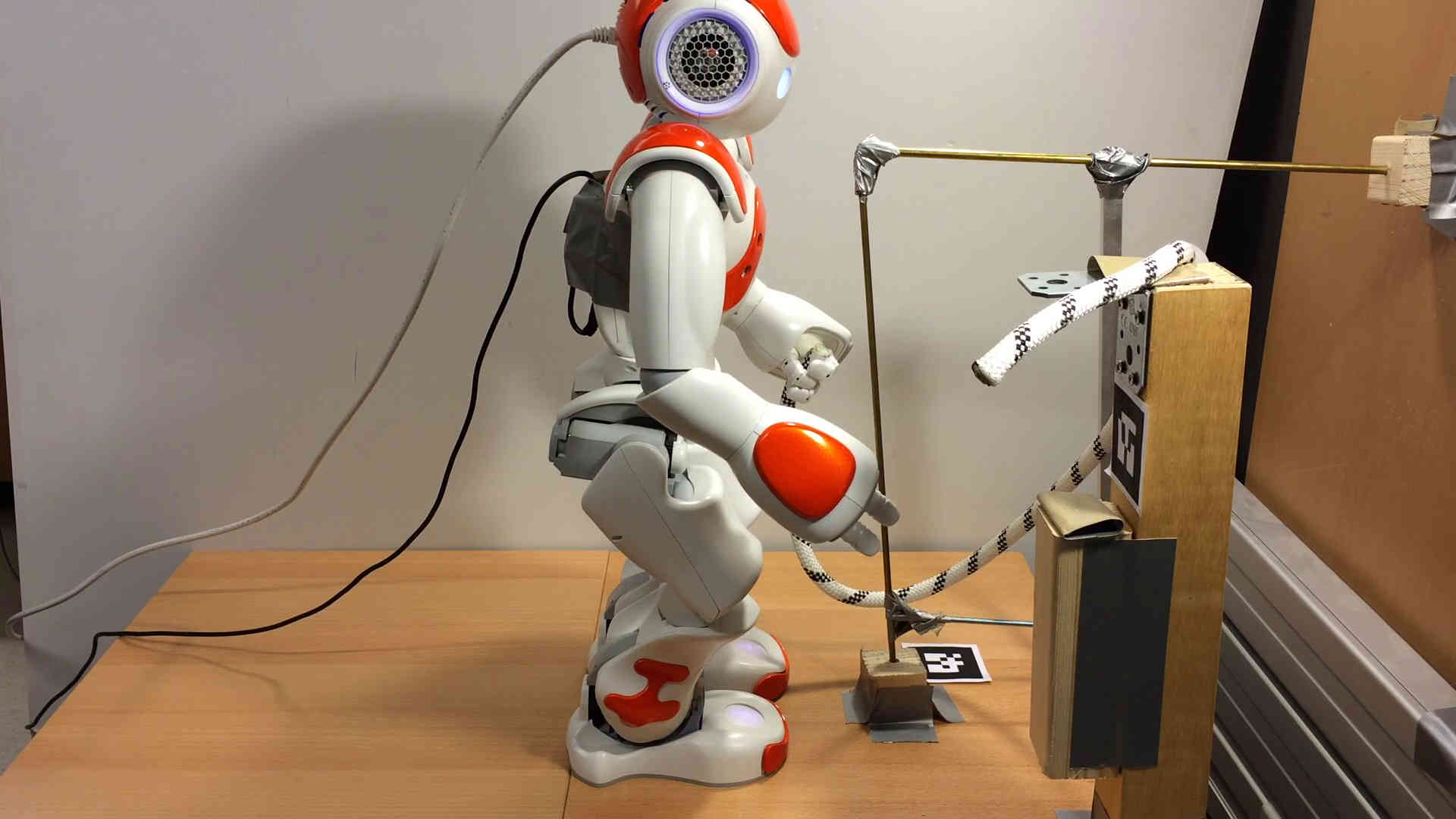}
\includegraphics[width = 0.15 \linewidth, trim={22cm 0cm 10cm 0cm}, clip]{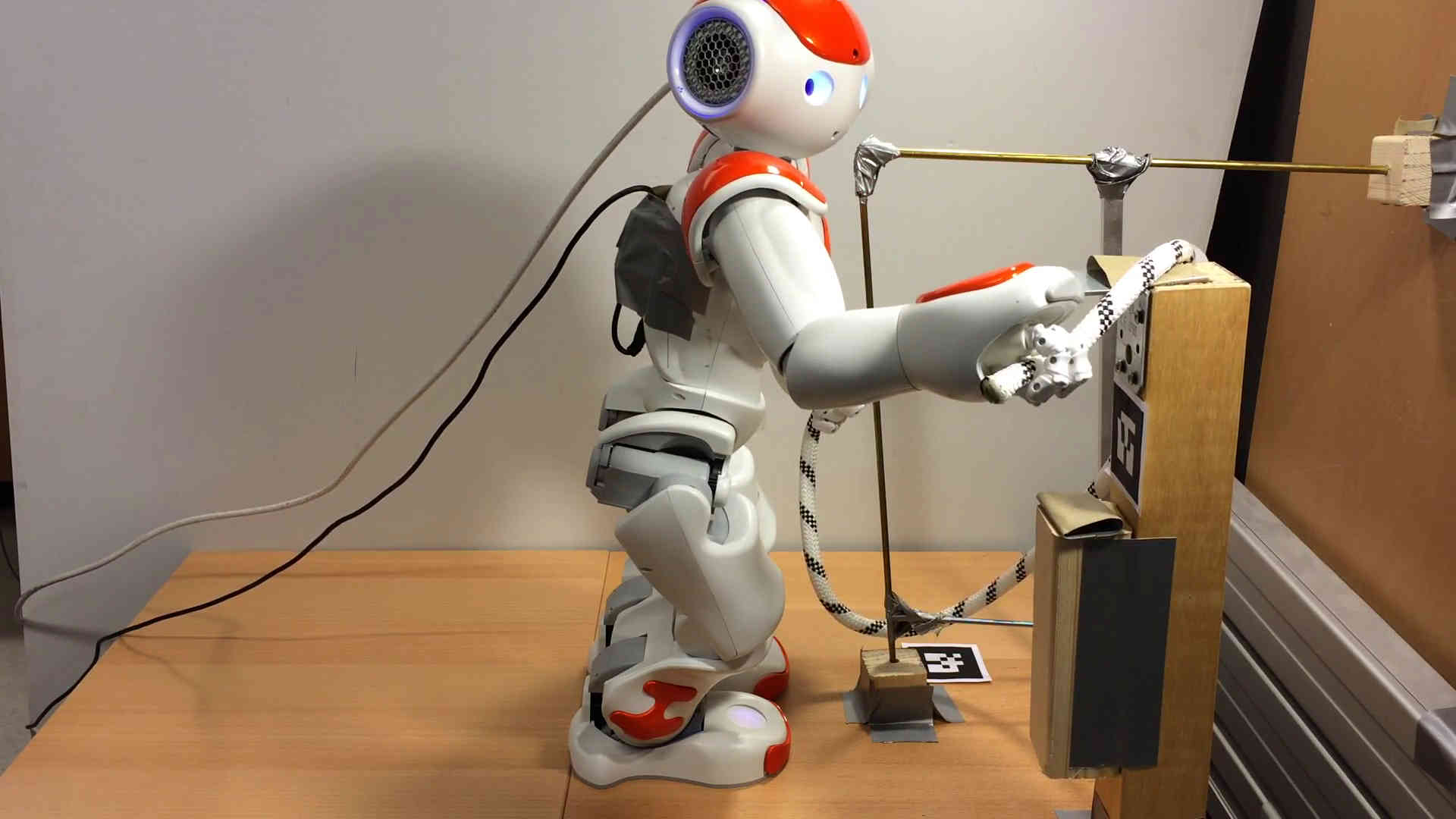}
\includegraphics[width = 0.15 \linewidth, trim={22cm 0cm 10cm 0cm}, clip]{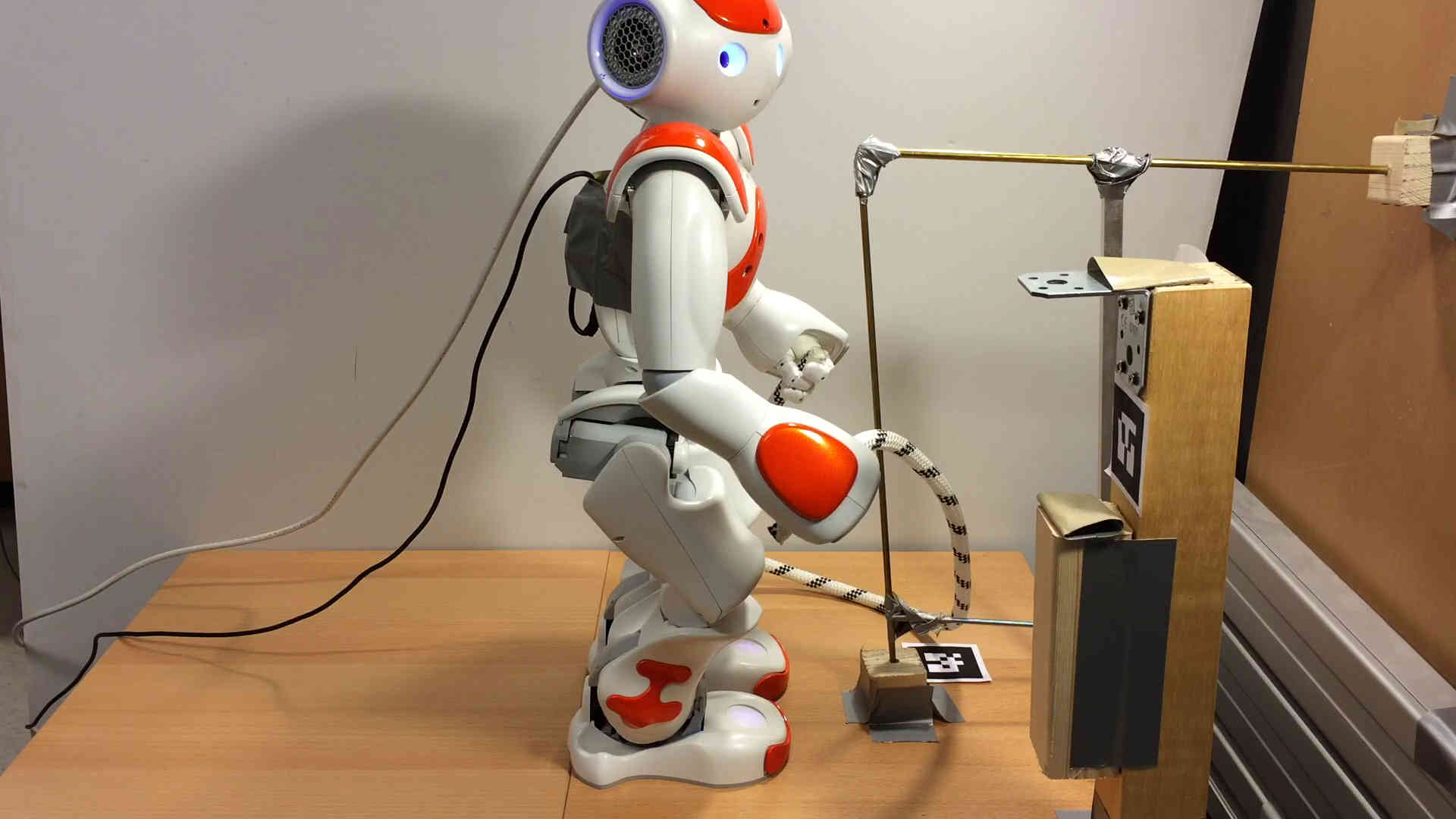}
\caption{NAO humanoid robot knotting the unknot. For this, the robot has to perform one insertion action. The unknot is finished when both ends of the rope are held and the rope passes through the anchoring entity.}
\label{fig:nao-unknot}

\includegraphics[width = 0.15 \linewidth, trim={22cm 0cm 10cm 0cm}, clip]{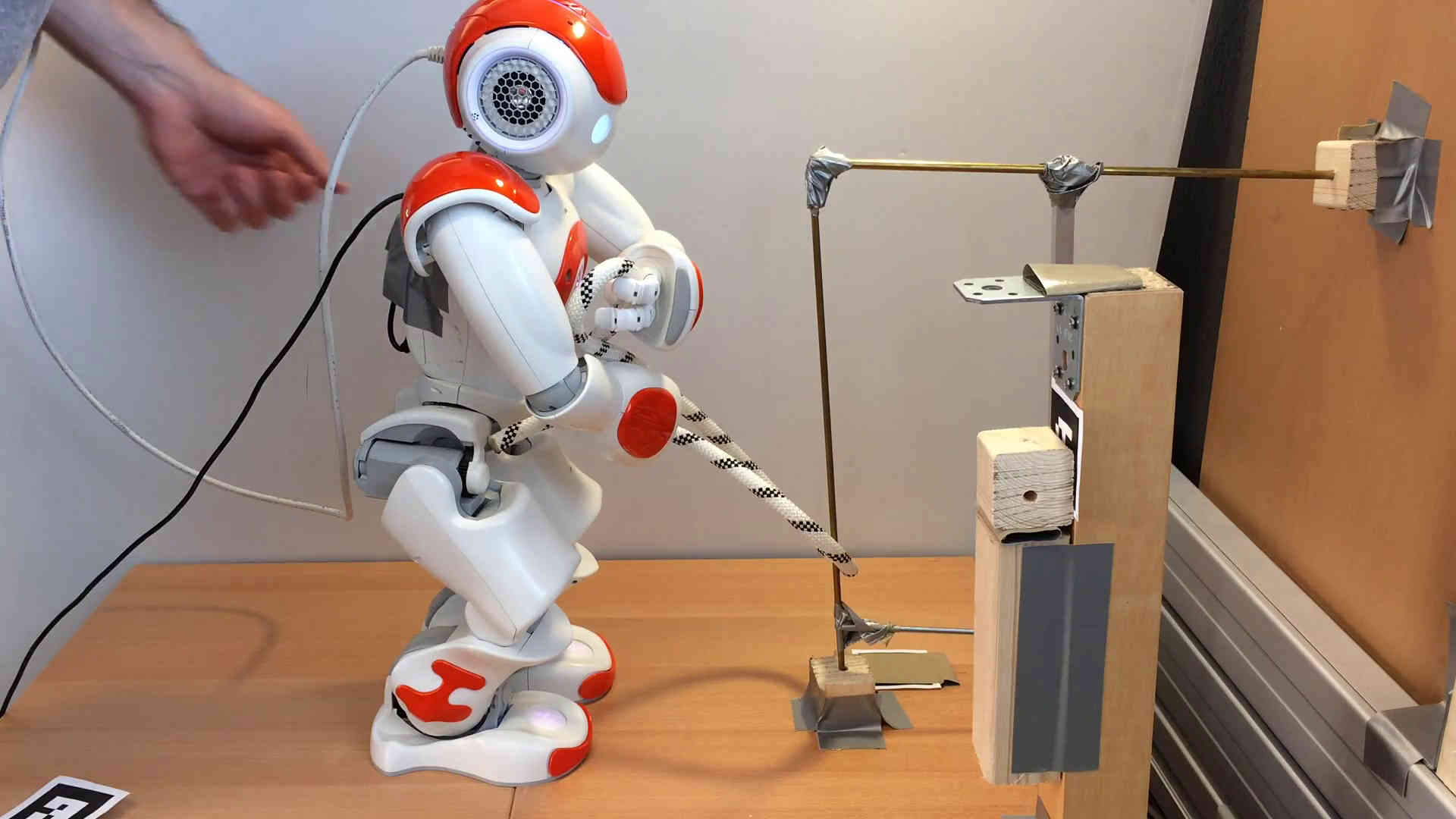}
\includegraphics[width = 0.15 \linewidth, trim={22cm 0cm 10cm 0cm}, clip]{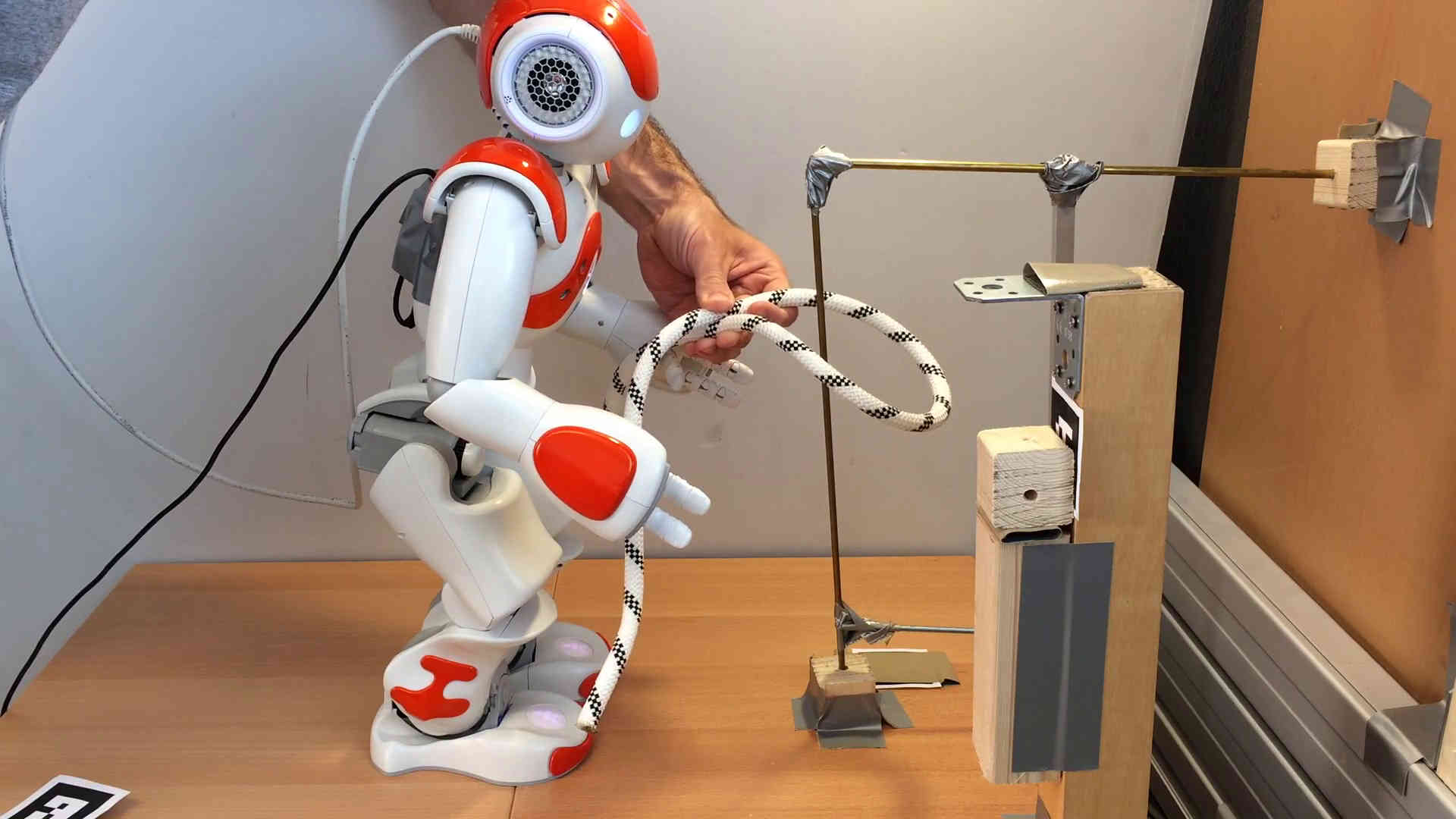}
\includegraphics[width = 0.15 \linewidth, trim={22cm 0cm 10cm 0cm}, clip]{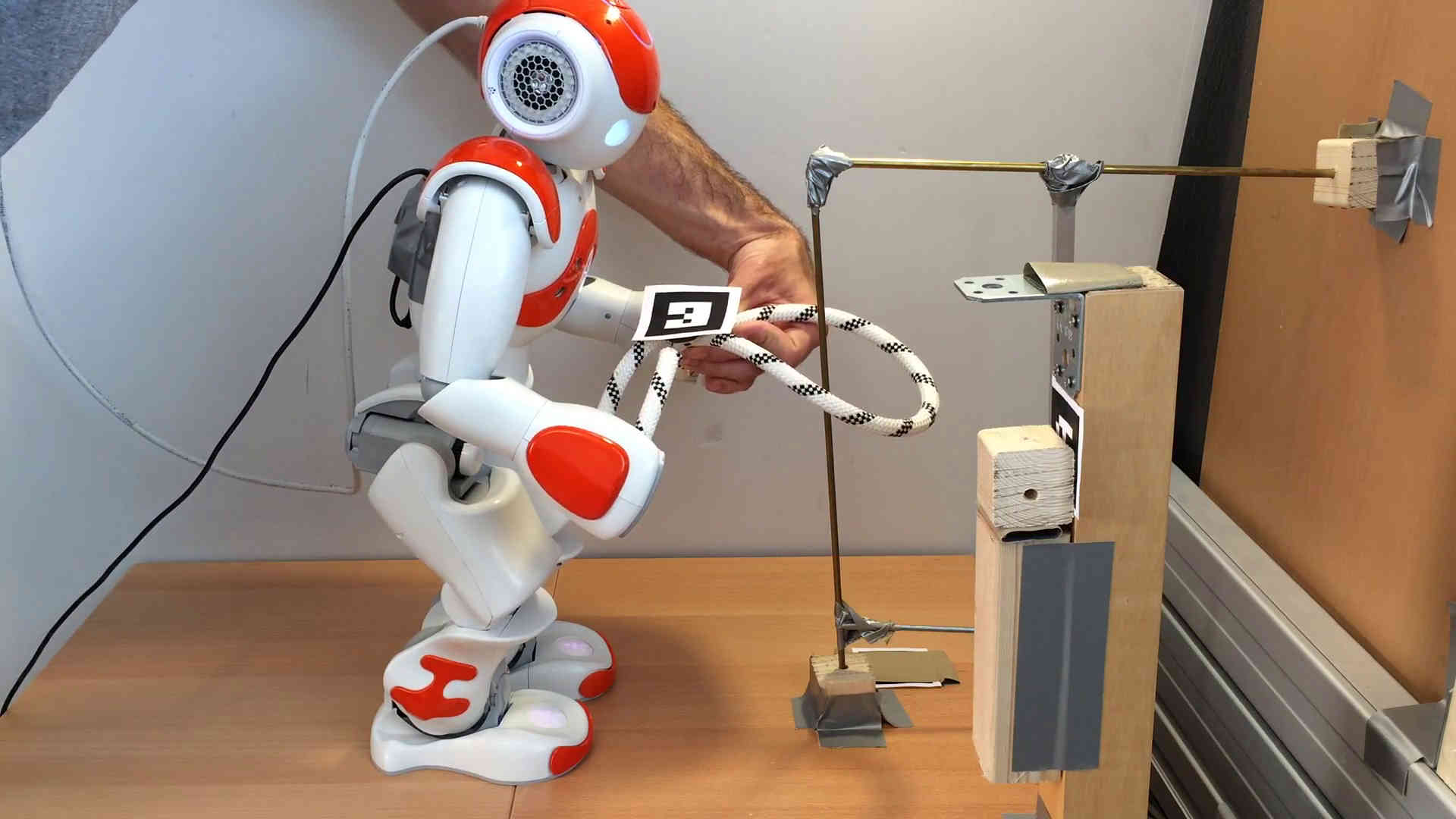}
\includegraphics[width = 0.15 \linewidth, trim={22cm 0cm 10cm 0cm}, clip]{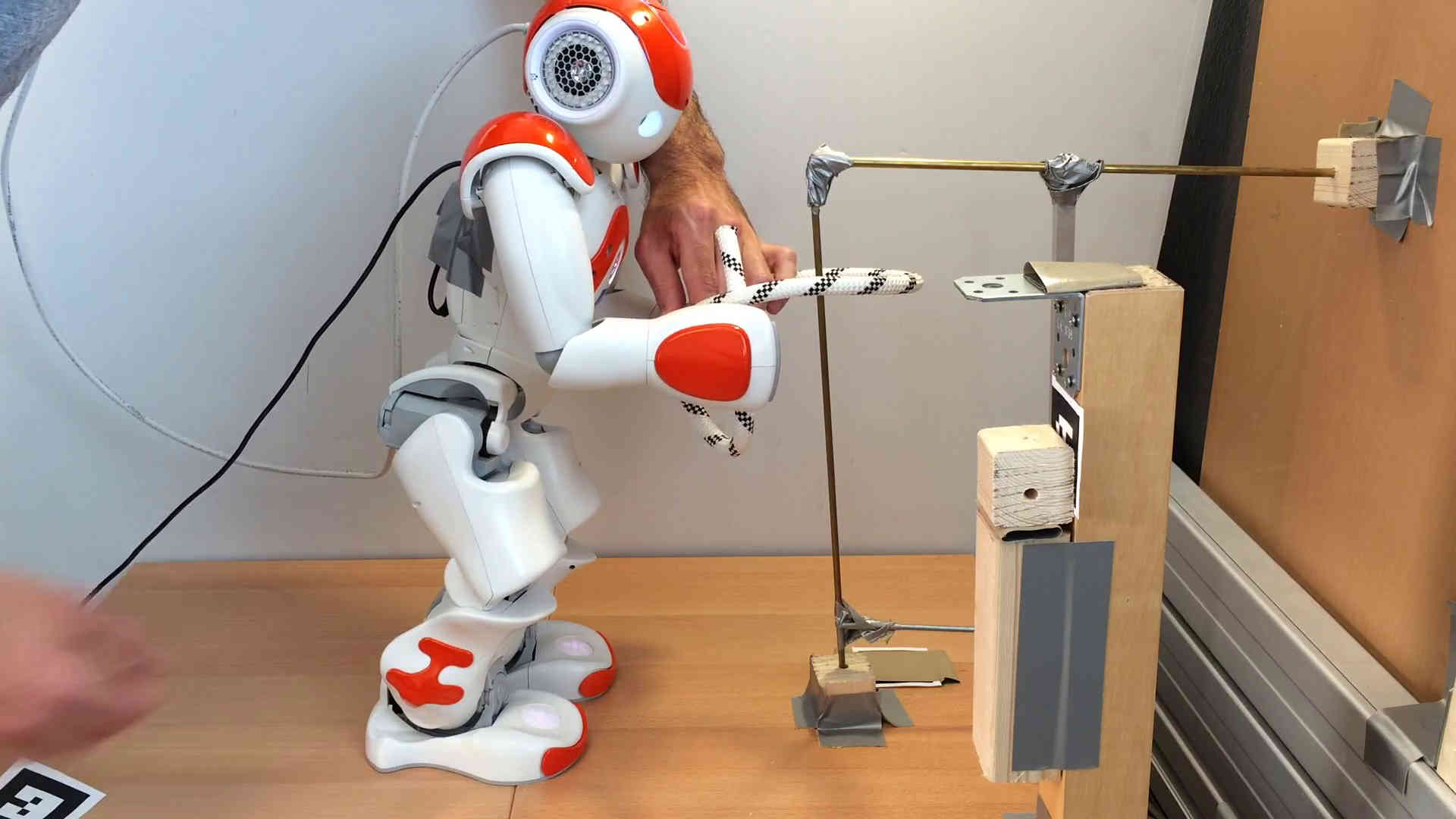}
\includegraphics[width = 0.15 \linewidth, trim={22cm 0cm 10cm 0cm}, clip]{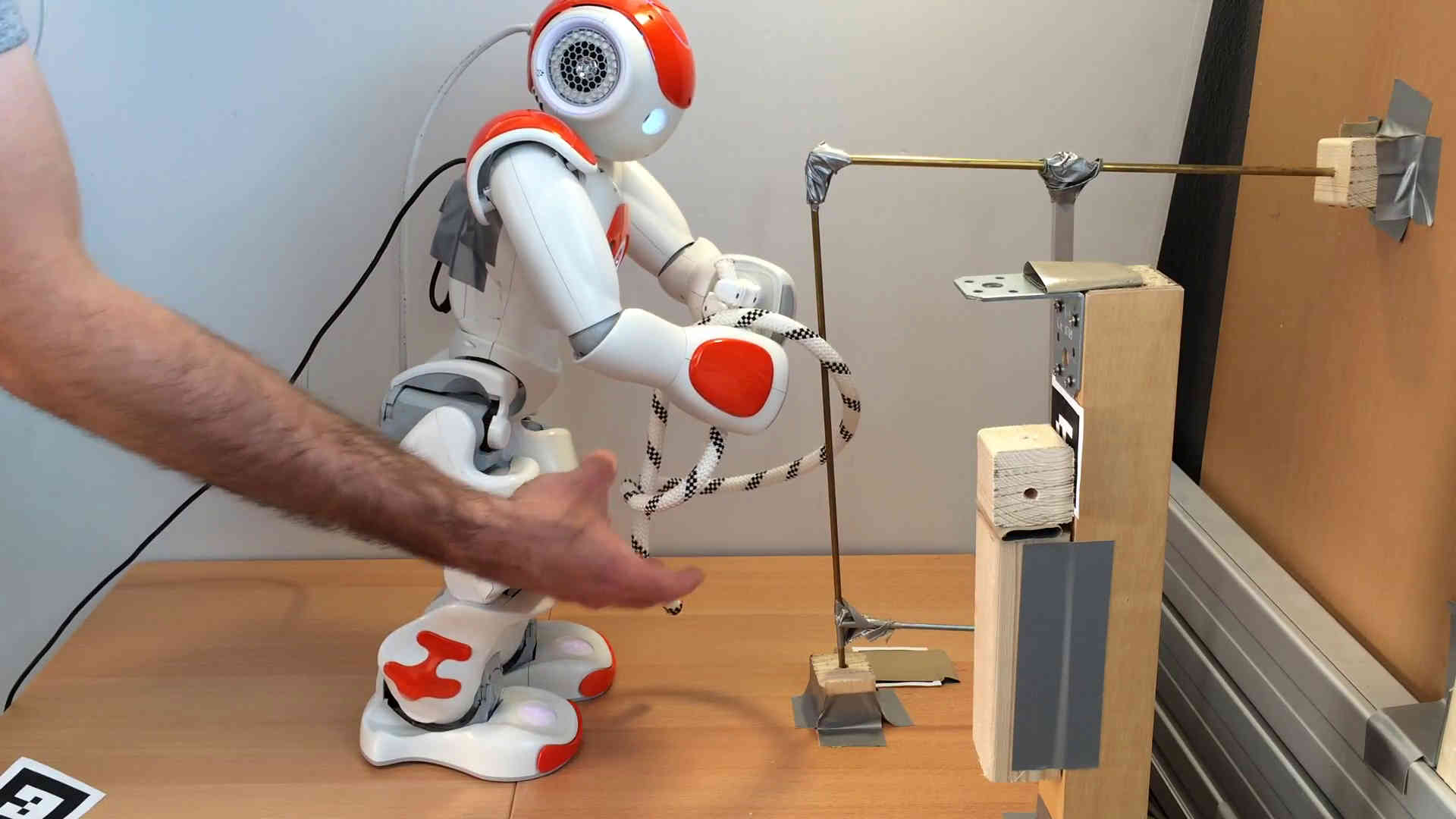}
\includegraphics[width = 0.15 \linewidth, trim={22cm 0cm 10cm 0cm}, clip]{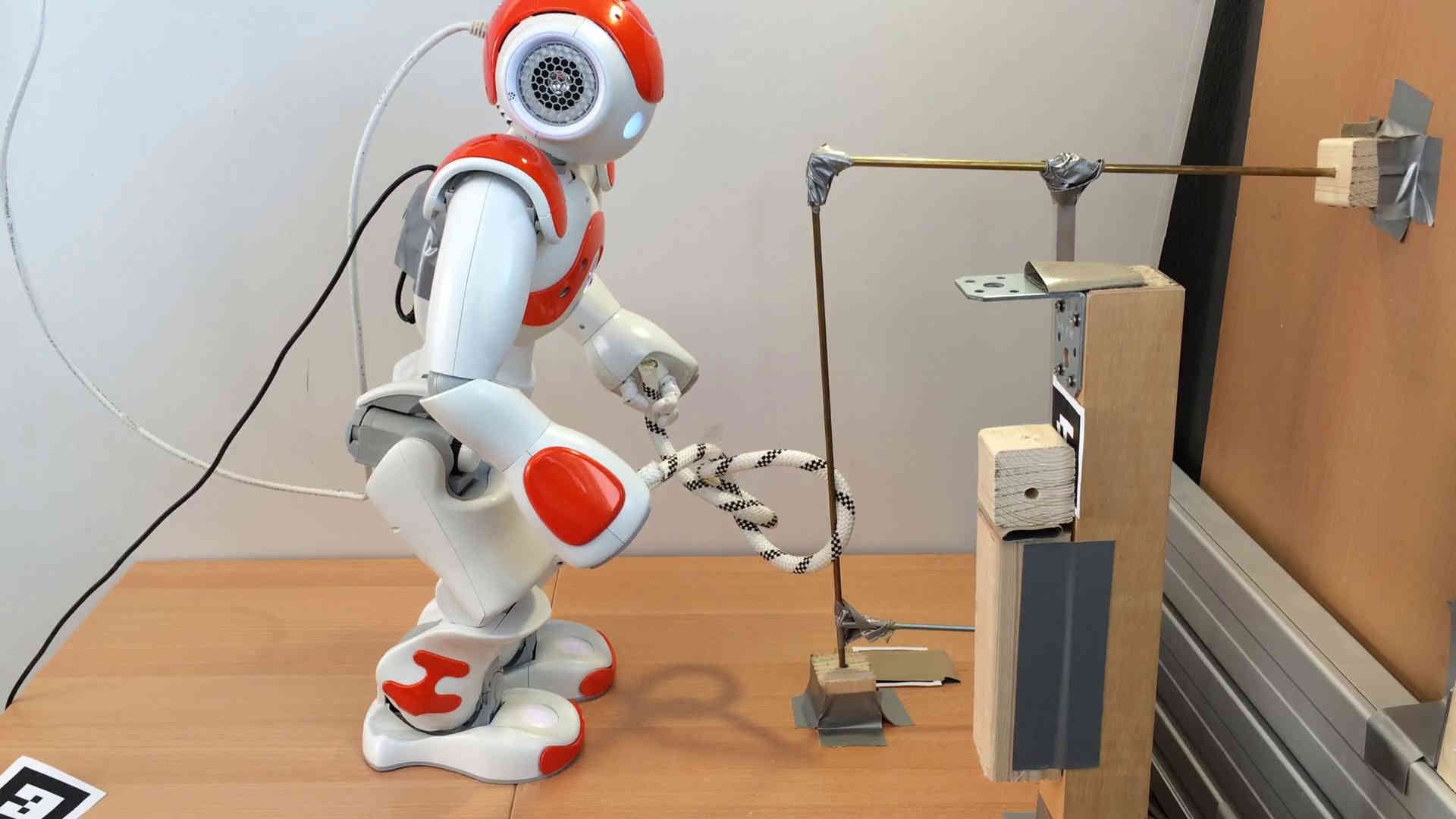}
\caption{NAO humanoid robot knotting the trefoil knot. For this, the robot continues from the unknot in Fig.~\ref{fig:nao-unknot} and has to perform one twisting action, one insertion action, and finally grasp the free end of the rope.}
\label{fig:nao-trefoil}
\end{figure*}

In this experiment, we test if the robot can execute complete sequences of basic knotting actions to create knots. First, we let the robot create the unknot around the anchoring entity and then continue with a trefoil knot. The process is documented in Fig.~\ref{fig:nao-unknot} and Fig.~\ref{fig:nao-trefoil}, read from left to right. Real-time videos are provided in the supplementary material. 

The experiments showed that the defined knotting actions suffice to control the robot accomplishing the knotting task successfully and robustly. It is clear that the robot's perception and manipulation abilities are too limited to tie complex knots in a completely autonomous fashion. Therefore, the robot requires assistance in holding the crossing point for the trefoil and regrasping the rope once released. The perception problem lies beyond the scope of our current work and it is therefore simplified for this demonstration.

\section{Conclusions}
\label{sec:conclusions}

We have presented and evaluated a method for the insertion action that recurrently appears in robotic knotting scenarios and showed how it can be used in conjunction with other basic knotting actions. Our magnetic field representation of the insertion action applies to target loops of all shapes and allows adjusting the quality and the speed of the insertion action through our parametrization. The reliability of our approach was validated in simulation experiments including dynamic loop deformation and motion. Further, we demonstrated how basic knotting actions can be used to create several knots by scheduling them in different sequences and described how BTs can represent redundancy and fallback handling which are inherent to knotting tasks. Lastly, we demonstrated the robustness of our insertion action by showcasing it with a NAO in a knotting task.

To the best of our knowledge, this is the first work proposing parametrized magnetic fields to address manipulation of deformable bodies in the context of knotting. By using magnetic fields we only rely on the loop coordinates and bypass the need for complex motion planners. This enables us to construct knots in responsive real time process, relying on a robust insertion action.

\section*{Acknowledgment}
This work has been supported by the Swedish Research Council (VR) and the European Union Project RECONFIG, \url{www.reconfig.eu} (FP7-ICT-2011-9, Project Number: 600825) and FLEXBOT (FP7-ERC-279933). The authors gratefully acknowledge the support.

\flushend
\bibliographystyle{IEEEtran}
\bibliography{ref_personal}

\end{document}